\documentclass[10pt,twocolumn,letterpaper]{article}
\pdfoutput=1

\usepackage{iccv}
\usepackage{times}
\usepackage{epsfig}

\usepackage[pagebackref=true,breaklinks=true,letterpaper=true,colorlinks,bookmarks=false]{hyperref}

\usepackage{microtype}
\usepackage{graphicx}
\usepackage{subfigure}
\usepackage{booktabs} 

\usepackage[table]{xcolor}
\definecolor{lightblue}{rgb}{0.4, 0.6, 0.9}
\definecolor{lightred}{rgb}{0.9, 0.6, 0.4}
\definecolor{lightgreen}{rgb}{0.3, 0.7, 0.4}

\newcommand{\matthijs}[1]{{\color{blue}[\textbf{Matthijs}:#1]}}
\newcommand{\dmitry}[1]{#1}
\newcommand{\zeki}[1]{{\color{green}[\textbf{Zeki}:#1]}}
\newcommand{\yash}[1]{{\color{orange}[\textbf{Yash}:#1]}}

\usepackage{pifont}

\graphicspath{{./}{../}}

\usepackage{amsthm}
\usepackage{amsmath}
\usepackage{amsfonts}
\usepackage{soul}
\usepackage{algorithm2e}

\newcommand{\fig}[1]{Figure~\ref{fig:#1}}
\newcommand{\sect}[1]{Section~\ref{sect:#1}}
\newcommand{\tab}[1]{Table~\ref{tab:#1}}

\newcommand{\aka}{{a.k.a}.\@ }

\usepackage{enumitem}
\usepackage{hyperref}
\newcommand{\ourmethod}{\textsc{DeDrift}\xspace}

\begin{document}
 \iccvfinalcopy 

\def\iccvPaperID{3639} 
\def\httilde{\mbox{\tt\raisebox{-.5ex}{\symbol{126}}}}

\ificcvfinal\pagestyle{empty}\fi

\title{\ourmethod: Robust Similarity Search under Content Drift}

\author{Dmitry Baranchuk\thanks{The work is partially done during the internship at Meta AI}  \\  
Yandex Research\\
{\tt\small dbaranchuk@yandex-team.ru } 
\and
Matthijs Douze\\
Meta AI\\
{\tt\small matthijs@meta.com}
 \and
Yash Upadhyay\\
Meta AI\\
{\tt\small yashup@meta.com}
 \and
I. Zeki Yalniz\\
Meta AI\\
{\tt\small izy@meta.com}
}

\maketitle
\ificcvfinal\thispagestyle{empty}\fi

\begin{abstract}
The statistical distribution of content uploaded and searched on media sharing sites changes over time due to seasonal, sociological and technical factors.
We investigate the impact of this ``content drift'' for large-scale similarity search tools, based on nearest neighbor search in embedding space.
Unless a costly index reconstruction is performed frequently, content drift degrades the search accuracy and efficiency. 
The degradation is especially severe since, in general, both the query and database  distributions change.

We introduce and analyze real-world image and video datasets for which temporal information is available over a long time period. 
Based on the learnings, we devise \ourmethod, a method that updates embedding quantizers to continuously adapt large-scale indexing structures on-the-fly. 
\ourmethod almost eliminates the accuracy degradation due to the query and database content drift while being up to $100{\times}$ faster than a full index reconstruction.

\end{abstract}


\section{Introduction}
\label{sect:intro}

\begin{figure*}
    \centering
    \vspace{-4mm}
    \includegraphics[width=0.99\linewidth]{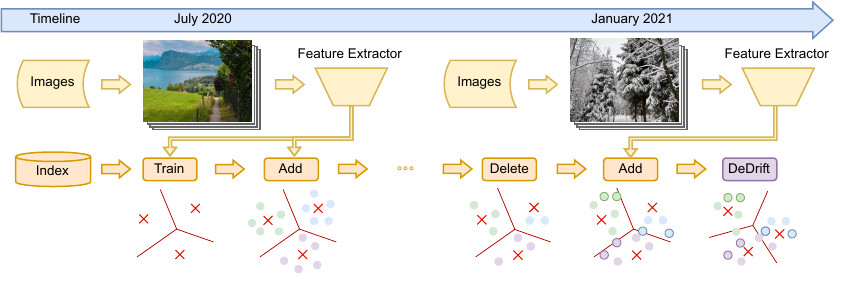}
    \caption{
        Overview of the dynamic index developed in this work. 
        Images are continuously uploaded to a online sharing platform and their embeddings are added to an index. 
        At any moment, the index can be used to search for similar images.
        The index quantizes the embeddings into centroids. 
        However, as the content drifts over time, the centroids do not match the data distribution anymore. 
        \ourmethod introduces a lightweight update procedure to adapt to the new data distribution. 
    }
    \label{fig:splash}
\end{figure*}

The amount of content available online is growing exponentially over the years. 
Various online content sharing sites collect billions to trillions of images, videos and posts over time. 
Efficient Nearest Neighbor Search (NNS) techniques enable searching these vast unstructured databases based on content similarity. 
NNS is at the core of a plethora of practical machine learning applications including 
content moderation ~\cite{douze2021isc}, 
retrieval augmented modeling  ~\cite{borgeaud2022improving,izacard2022few},
keypoint matching for 3D reconstruction~\cite{agarwal2011building},
image-based geo-localisation~\cite{chen2011city},
content de-duplication~\cite{wu2007ccweb}, 
k-NN classification~\cite{assran2021semi,Caron_2021_ICCV}, 
defending against adversarial attacks ~\cite{DubeyCVPR19}, 
active learning ~\cite{ColemanAAAI22} 
and many others.

NNS techniques extract high dimensional feature vectors (\aka ``embeddings'') from each item to be indexed. These embeddings may be computed by hand-crafted techniques ~\cite{pls+10,jegou2012aggregating} or, more commonly nowadays, with pre-trained neural networks ~\cite{babenko2014neural,radenovic2018fine,pizzi2022sscd}.
Given the database of embedding vectors $\mathcal{D}{=}\{x_1,\dots,x_N\} \subset \mathbb{R}^d$ and a query embedding $q \in \mathbb{R}^d$, NNS retrieves the closest database example to $q$ from $\mathcal{D}$ according to some similarity measure, typically the L2 distance: 
\begin{equation}
\mathrm{argmin}_{x\in \mathcal{D}} \| q - x \|^2
\end{equation} 
The exact nearest neighbor and its distance can be computed by exhaustively comparing the query embeddings against the entire database.
On a single core of a typical server from 2023, this brute-force solution takes a few milliseconds for databases smaller than few tens of thousand examples with embeddings up to a few hundred dimensions.  


However, practical use cases require real time search on millions to trillion size databases, where brute-force NNS is too slow~\cite{chern2023tpuknn}.
Practitioners resort to approximate NNS (ANNS), trading some accuracy of the results to speed up the search.
A common approach is to perform a statistical analysis of $\mathcal{D}$ to build a specialized data structure (an ``index'' in database terms) that performs the search efficiently. 
Like any unsupervised machine learning task, the index is trained on representative sample vectors from the data distribution to the index. 

A tool commonly used in indexes is vector quantization~\cite{gray1998quantization}. 
It consists in representing each vector by the nearest vector taken in a finite set of centroids $\{c_1,\dots,c_K\}\subset \mathbb{R}^d$. 
A basic use of quantization is compact storage: the vector can be reduced to the index of the centroid it is assigned to, which occupies just $\lceil \log_2 K \rceil$ bits of storage. 
The larger $K$, the better the approximation. 
As typical for machine learning, the training set $\mathcal{T}$ is distinct from $\mathcal{D}$ (usually $\mathcal{T}$ is a small subset of $\mathcal{D}$). 
The unsupervised training algorithm of choice for quantization is k-means which guarantees Lloyd's optimality conditions~\cite{lloyd1982least}.

In an online media sharing site, the index functions as a database system. 
After the initial training, the index ingests new embeddings by batches as they are uploaded by users and removes content that has been deleted (Figure~\ref{fig:splash}). 
Depending on the particular indexing structure, addition/deletion may be easy~\cite{Jegou11a}, or more difficult~\cite{Singh2021freshdiskann}. 
However, a more pernicious problem that appears over the time is content drift. 
In practice, over months and years, the statistical distribution of the content changes slowly, both for the data inserted in the index and the items that are queried.

\begin{figure}
\centering
\begin{tabular}{cccccc}
\hspace{-3mm}
\includegraphics[width=\linewidth]{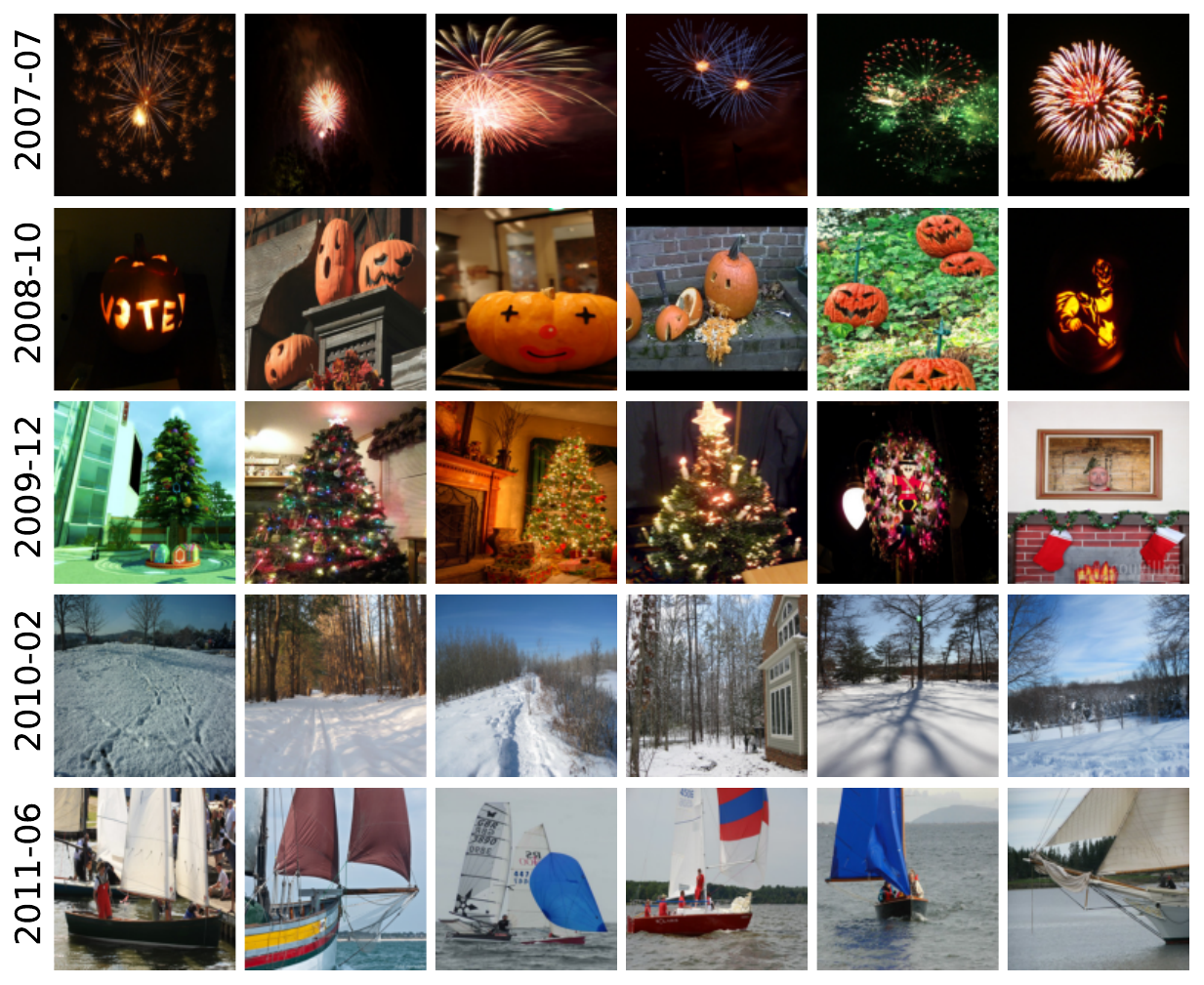}%
\rotatebox{90}{\tiny Image credits: see Appendix~\ref{app:imcredits}} 
\end{tabular}
\caption{
	Per row: sample YFCC images that are typical for some months. 
	Refer to~\ref{sec:permonthsim} for how the images were selected. 
}
\label{fig:visialize_yfcc}
\end{figure}

The drift on image content may have a technical origin, \eg 
camera images become sharper and have better contrast; 
post-processing tools evolve as many platforms offer editing options with image filters that are applied to millions of images. 
The drift may be seasonal: the type of photos that are taken in summer is not the same as in winter, see Figure~\ref{fig:visialize_yfcc}.
Sociological causes could be that people post pictures of leaderboards of a game that became viral, or there is a lockdown period where people share only indoor pictures without big crowds. 
In rare cases, the distribution may also change suddenly, for example because of an update of the platform that changes how missing images are displayed.
The problem posed by this distribution drift is that new incoming vectors are distributed differently from $\mathcal{T}$. 
Indeed, by design, feature extractors are sensitive to semantic differences in the content. 
This mismatch between training and indexed vectors impacts the search accuracy negatively. 
To address this, practitioners monitor the indexing performance and initiate a full index reconstruction once the efficiency degrades noticeably. 
By definition, this is the optimal update strategy since it adapts exactly to the new data distribution that contains both old and recent vectors. 
However, at larger scales, this operation becomes a resource bottleneck since all $N$ vectors need to be re-added to the index, and disrupts the services that relies on the index. 

Our first contribution is to carefully investigate temporal distribution drift in the context of large-scale nearest neighbor search. 
For this purpose, we introduce two real-world datasets that exhibit drift. 
We first measure the drift in an index-agnostic way, on exact search queries (Section~\ref{sec:stats}). 
Then we measure the impact on various index types that are commonly used for large-scale vector search (Section~\ref{sec:indexdrift}).

Our second contribution is \ourmethod, a family of adaptation strategies applied to the most vulnerable index types (Section~\ref{sec:indexupdate}). 
\ourmethod modifies the index slightly to adapt to the evolution of the vector distribution, without re-indexing all the stored elements, which would incur an $\mathcal{O}(N)$ cost. 
This adaptation yields search results that are close to the reindexing topline while being  orders of magnitude faster. 
This modification is done while carefully controlling the accuracy degradation. 
Sections~\ref{sec:results} reports and analyzes \ourmethod's results. 

\section{Related work}
\label{sect:related}

\noindent 
{\bf NNS methods.}
In low-dimensional spaces, NNS can be solved efficiently and exactly with tree structures like the KD-tree~\cite{friedman1977algorithm,muja2014scalable} and ball-trees~\cite{omohundro1989five}, that aim at achieving a search time logarithmic in $N$. 
However, in higher dimensions, due to the \textit{curse of dimensionality}, the tree structures are ineffective to prune the search space, so there is no efficient exact solution. 
Therefore, practitioners use approximate NNS (ANNS) methods, trading some accuracy to improve the efficiency.
Early ANNS methods rely on data-independent structures, \eg projections on fixed random directions~\cite{andoni2005locality,datar2004locality}. 
However, the most efficient methods adapt to the data distribution using vector quantization. 

The Inverted File (IVF) based indexing relies on a vector quantizer (the \emph{coarse quantizer}) to partition the database vectors into clusters~\cite{sivic2003video,Jegou11a}.
At search time, only one or a few of these clusters need to be checked for result vectors. 
Such pruning approach is required to search in large datasets, so we focus on IVF-based methods in this paper.  
When scaling up the coarse quantizer can become the computation bottleneck. 
Several alternatives to plain k-means have been proposed: 
the inverted multi-index uses a product quantizer~\cite{Babenko12}, 
graph-based indexes~\cite{Baranchuk_2018_ECCV} or residual quantizers can also be used~\cite{RVQ10}.

Out-of-distribution (OOD) vectors w.r.t the training distribution are detrimental to the search performance~\cite{simhadri22results}. 
For IVF based methods, they translate to unbalanced cluster sizes, which incurs a slowdown that can be quantified by an \emph{imbalance factor}~\cite{jegou2010improving}. 
In~\cite{zhang2020continuously}, OOD content is addressed before indexing with LSH by adapting a transformation matrix, but LSH's accuracy is limited by its low adaptability to the data distribution. 
Graph-based pruning methods like HNSW~\cite{HNSW} also suffer from OOD queries~\cite{jaiswal2022ood}. 
In this work, our focus is not on graph-based indexes since they do not scale as well to large datasets.

\noindent 
{\bf Database drift.}
In the literature, ANNS studies are primarily on publicly available offline vector databases where the distribution of query and database examples are fixed and often sampled from the same underlying data distribution \cite{jegou2011searching,babenko2016efficient}. 
However, these assumptions do not typically hold for  real world applications. 
Not only the query frequency and database size, but also their distributions may drift over time. 
\dmitry{Recent works~\cite{onlinepq, onlineaq} simulate this setting and propose adaptive vector encoding methods. 
In our work, we collect the data with natural content drift and observe that  
vector codecs are relatively robust to content changes, as opposed to IVF-based indexing structures.
Another work~\cite{evolvekmeans} adapts the k-means algorithm to take the drift into account. 
Though this method can improve the performance of IVF indexes, it still requires full index reconstruction.}




\begin{figure}[t]
\begin{tabular}{cc}
\multicolumn{2}{c}{VideoAds} \vspace{-1mm} \\

\hspace{-3mm}\includegraphics[width=0.7\linewidth]{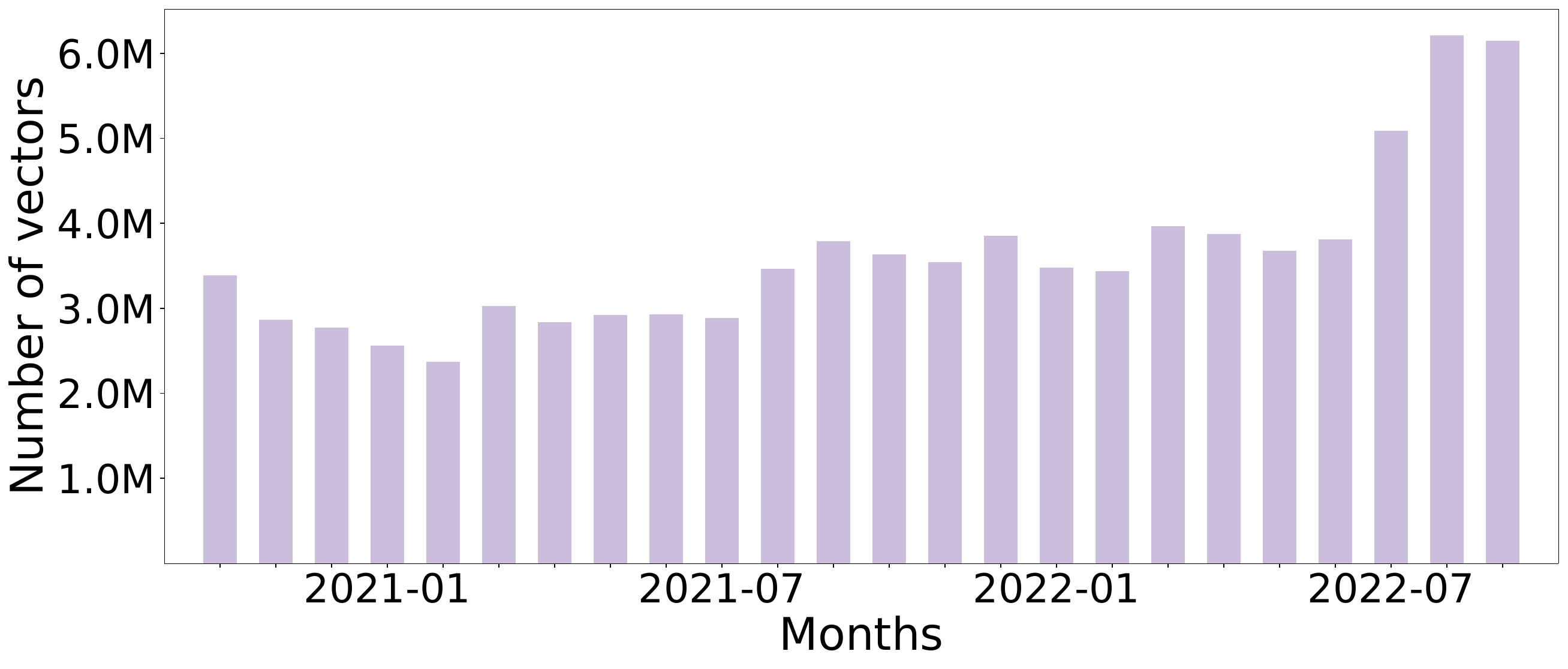} &
\hspace{-4mm}\includegraphics[width=0.32\linewidth,trim=0 -1 0 0,clip]
{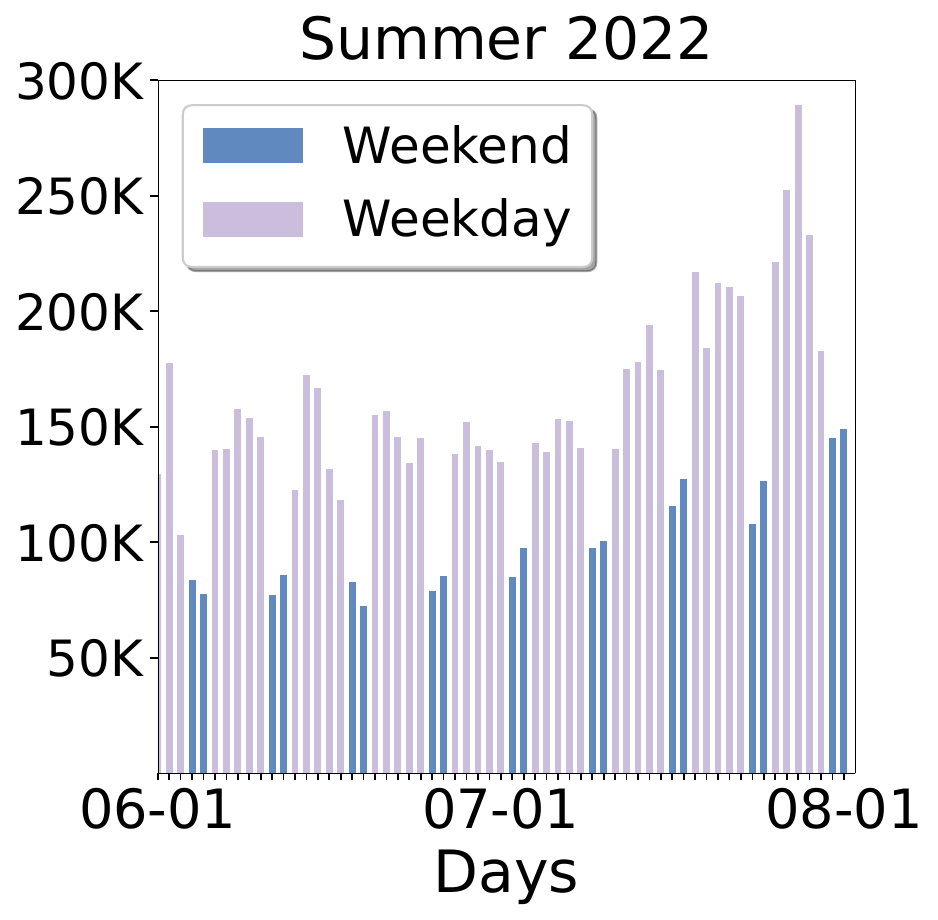} \\
\multicolumn{2}{c}{YFCC} \vspace{-1mm} \\
\hspace{-3mm}\includegraphics[width=0.7\linewidth]{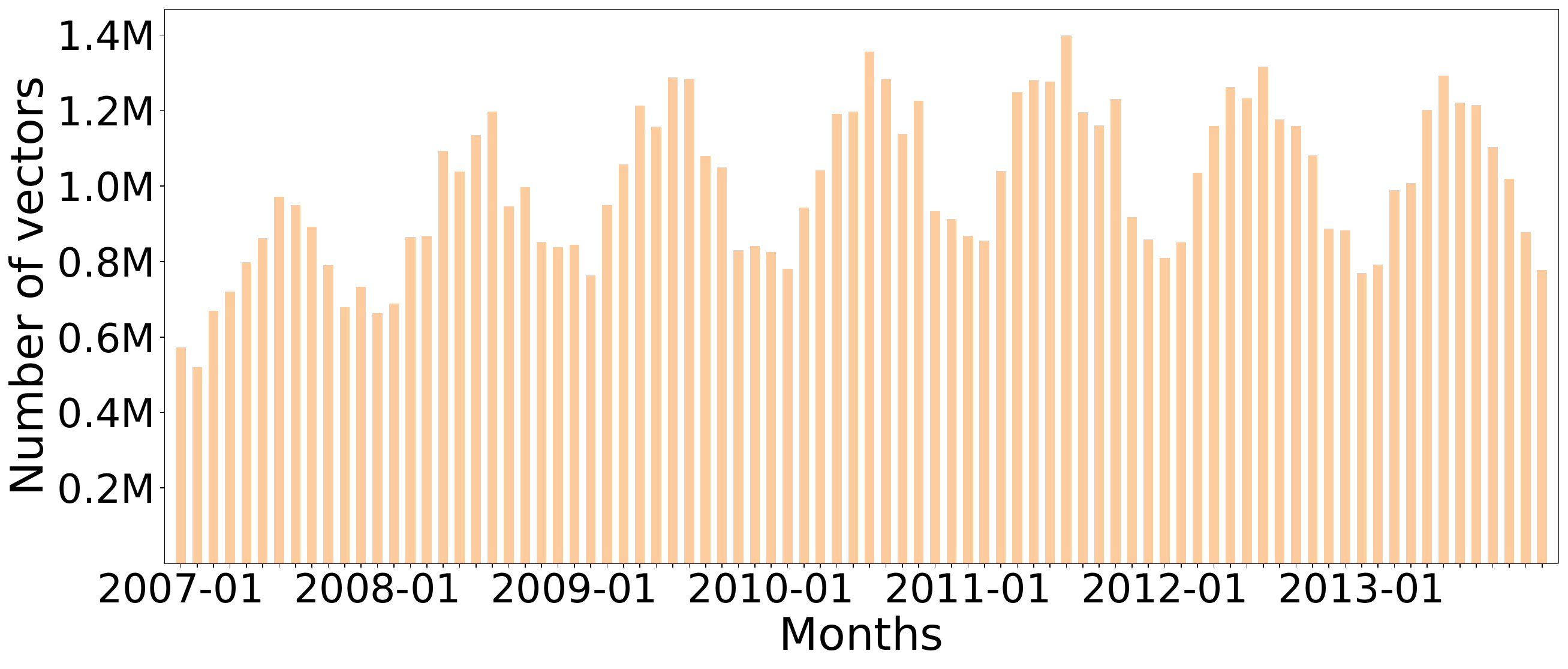} &
\hspace{-4mm}\includegraphics[width=0.32\linewidth,trim=0 -1 0 0,clip]
{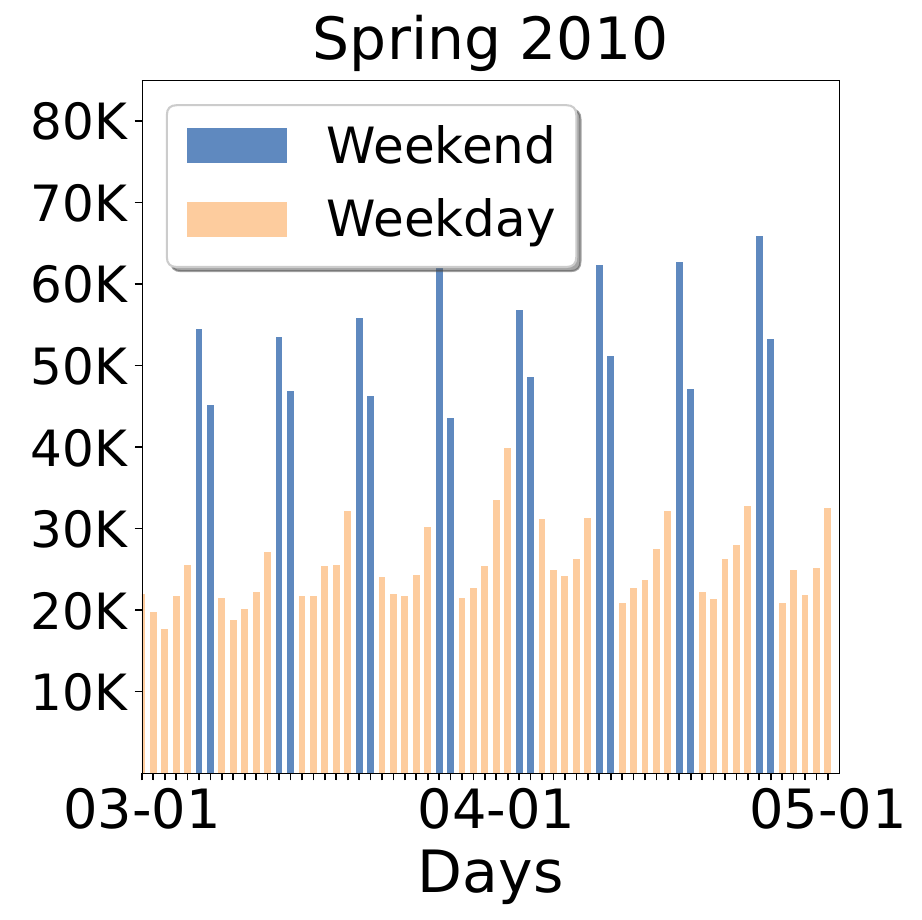} \\
\end{tabular}
\caption{
	Number of vectors per month (left) and per day on a smaller time span (right). 
}
\label{fig:dataset_sizes}
\end{figure}

\section{Temporal statistics of user generated content}
\label{sec:stats} 

We present the two representative datasets and report some statistics about them. 

\subsection{Datasets}

We introduce two  real-world datasets to serve as a basis for the drift analysis.
In principle, most content collected over several years exhibits some form of drift. 
For the dataset collection, the constraints are 
(1) the content should be collected over a long time period 
(2) the volume should be large enough that statistical trends become apparent
(3) the data source should be homogeneous so that other sources of noise are minimized 
(4) of course, timestamps should be available, either for the data acquisition or the moment the data is uploaded. 


The {\bf VideoAds} dataset contains ``semantic''\footnote{We call ``semantic'' embeddings that are intended for input to classifiers. 
In contrast, copy detection embeddings identify lower-level features useful for image matching~\cite{pizzi2022sscd}. } embeddings extracted from ads videos published on a large social media website between October $2020$ and October $2022$. 
The ads are of professional and non-professional quality, published by individuals, small and large businesses. 
They were clipped to be of maximum $30$ seconds in length. 
The video encoder for the video ads consisted of a RegNetY~\cite{RegNetY} frame encoder followed by a Linformer~\cite{Linformer} temporal pooling layer which were trained using MoCo~\cite{he2020momentum}, resulting in one embedding per video.
The dataset contains $86M$ L2-normalized embeddings in $512$ dimensions. 


The {\bf YFCC} dataset~\cite{Thomee2016YFCC100MTN} contains 100M images and videos from the Yahoo Flickr photo sharing site. 
Compared to typical user-generated content sites, Flickr images are of relatively good quality because uploads are limited and users are mostly pro or semi-pro photographers. 
We filter out videos, broken dates and dummy images of uniform color, leaving  $84M$ usable images spanning $7$ years, from January 2007 to December 2013. 
As a feature extractor, we use a pretrained DINO~\cite{Caron_2021_ICCV} model with ViT-S/16 backbone that outputs unnormalized 384-dimensional vectors. 
As DINO embeddings have been developed for use as k-NN classifiers, their L2 distances are semantically meaningful. 





Before evaluations, we apply a random rotation to all embeddings and uniformly quantize them to $8$ bits to make them less bulky.
These operations have almost no impact on neighborhood relations. 
We do not apply dimensionality reduction methods, e.g., PCA, since it can be considered as a part of the indexing method and explored separately. 

\subsection{Volume of data} 

First, we depict daily and monthly traffic for both datasets in \fig{dataset_sizes}.
For VideoAds, the traffic is lower during weekends and slightly increases over $2$ years. 
For YFCC, the traffic is higher on weekends than on weekdays. 
This pattern is due to the professional vs. personal nature of the data.

Over the years, the amount of content also grows for both datasets, following the organic growth of online traffic. 
Interestingly, for YFCC, we observe minimum traffic in January and maximum in July. 
In the following, we use subsampling to cancel the statistical effect of data volume.
Unless stated, we use a monthly granularity because this is the typical scale at which drift occurs.

\begin{figure}
\centering
\vspace{-2mm}
\begin{tabular}{cc}
\hspace{-2mm} VideoAds & \hspace{-4mm} YFCC  \\
\hspace{-2mm}{\footnotesize Oct 2020 $\rightarrow$ Sep 2022} & \hspace{-3mm}{\footnotesize Jan 2007 $\rightarrow$ Dec 2013}\\
\hspace{-3mm} \includegraphics[width=0.5\linewidth]{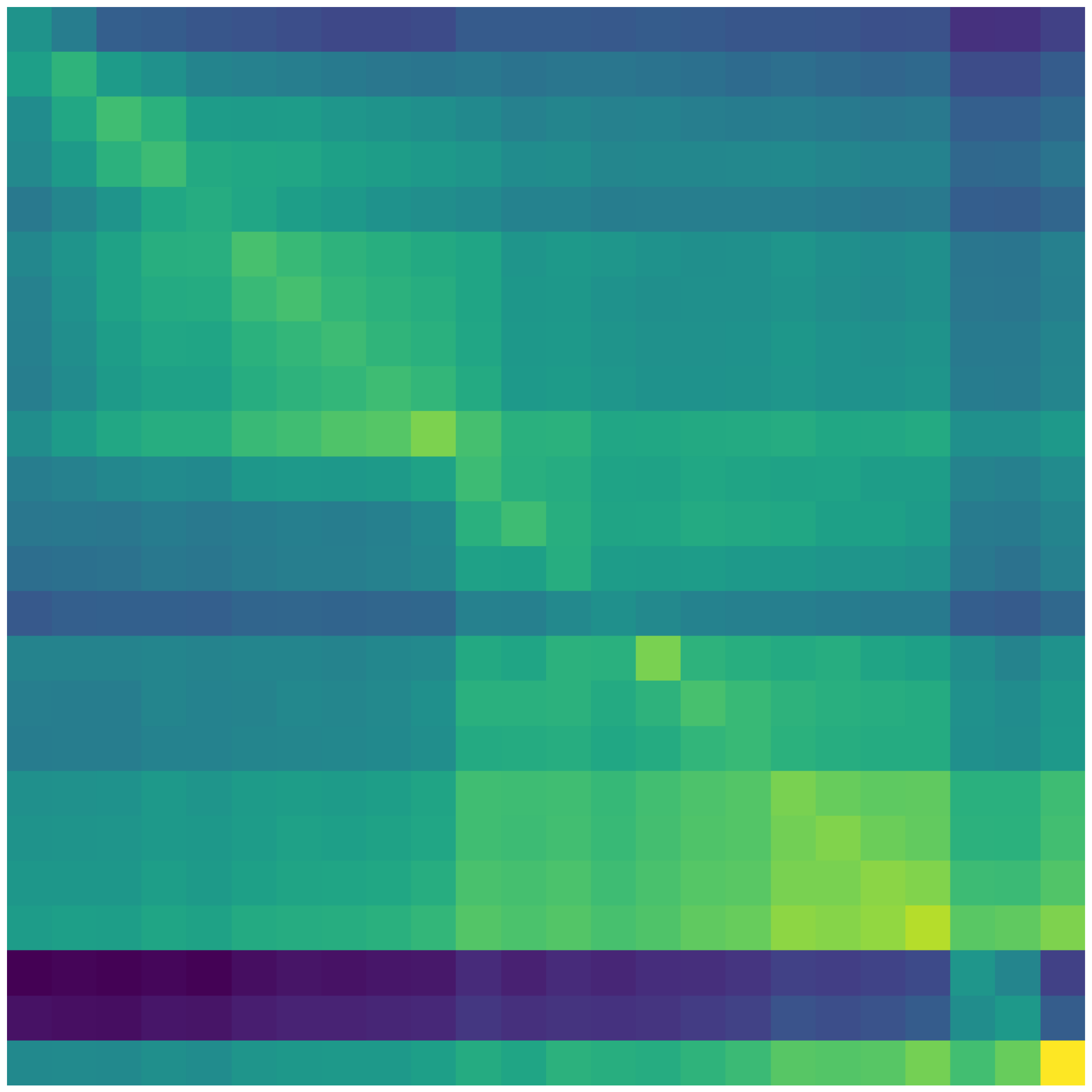} &
\hspace{-5mm}  \includegraphics[width=0.5\linewidth]{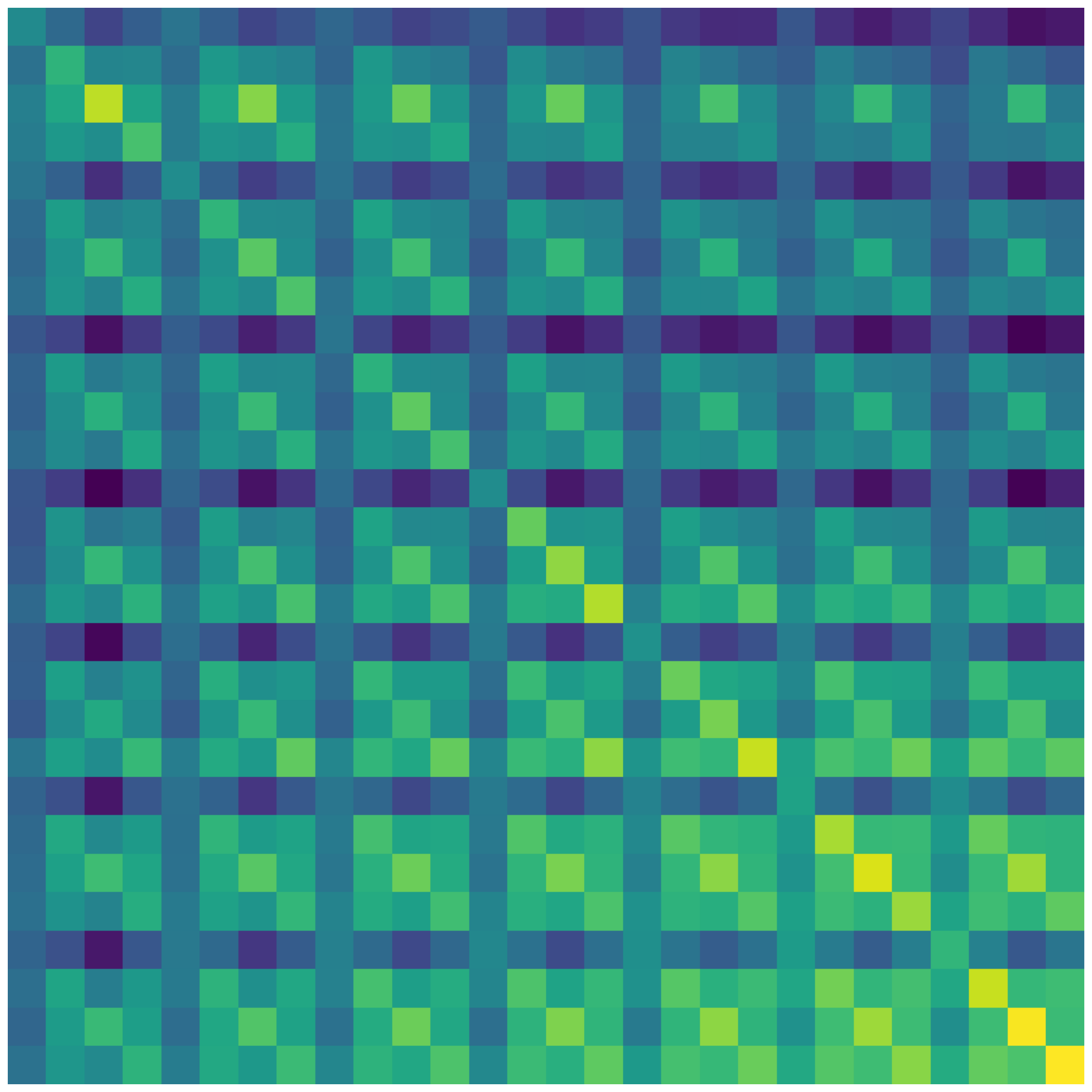} 
\end{tabular}
\caption{Pairwise similarity matrices between the embeddings over a time period subdivided in steps (1 month for VideoAds and 3 months for YFCC).
Blue and yellow correspond to low and high similarities, respectively. 
Both datasets have noticeable distribution drift over time. 
In addition, YFCC has clear seasonal correlations.
}
\vspace{-2mm}
\label{fig:similarity_matrices}
\end{figure}

\subsection{Per-month similarities} 
\label{sec:permonthsim}

We start the analysis of temporal distribution drift by visualizing pairwise similarities between months $i$ and $j$. 
For each month, we sample a random set $\Phi_{i} = \{\phi_i^1,..,\phi_i^n\}$ of $\lvert \Phi_{i} \rvert{=}10^5$  embeddings.
Then, we compute the similarity between $\Phi_i$ and $\Phi_j$ by averaging nearest-neighbor distances:
\begin{align}
    d(x, \Phi_j) = \frac{1}{L} \sum_{\ell=1}^{L} \|x - \mathrm{NN}_\ell(x, \Phi_{j})\|_2 \\
    S(\Phi_{i}, \Phi_{j}) = -\frac{1}{\lvert \Phi_{i} \rvert} \sum_{\phi_i \in \Phi_{i}} d(\phi_i, \Phi_{j}),
\end{align}
where $L{=}100$ and $\mathrm{NN}_\ell(x, \Phi)$ is the $\ell$-th nearest neighbor of $x$ in $\Phi$. 
Note that the similarity is asymmetric: $S(\Phi_i, \Phi_{j}){\neq}S(\Phi_{j}, \Phi_{i})$ in general. 
\fig{similarity_matrices} shows the similarity matrices for VideoAds and YFCC datasets. 
The analysis with daily and weekly granularities is in Appendix~\ref{app:simmat}. 

We observe that the ads data changes over time but does not have noticeable patterns. 
YFCC content drifts over years and has a strong seasonal dependency. 
\fig{visialize_yfcc} presents images from YFCC for different seasons\footnote{ 
The images for the month $i$ were selected as follows: a quantizer of size $K=4096$ is trained on $\Phi_0$. 
Vectors from $\Phi_i$ and $\Phi_{i-3}$ are assigned to the $K$ centroids independently.
We visualize six random images from one of the top-8 clusters where the size ratio between assignment $i$ over assignment $i{-}3$ is the largest. 
Out of these 8, we select one representative cluster for visualization.
}
the seasonal pattern is caused by images of outdoor sceneries (e.g., snow and foliage), seasonal activities (e.g., yachting) or holidays (e.g., Halloween and Christmas). 
Appendix~\ref{app:clusterbalance} studies the impact of these monthly similarity patterns on the distribution of quantization assignments.

\begin{figure}
\vspace{-2mm}
\begin{tabular}{cc}
\hspace{15mm} VideoAds & \hspace{23mm} YFCC
\end{tabular}\\
\includegraphics[width=\linewidth]{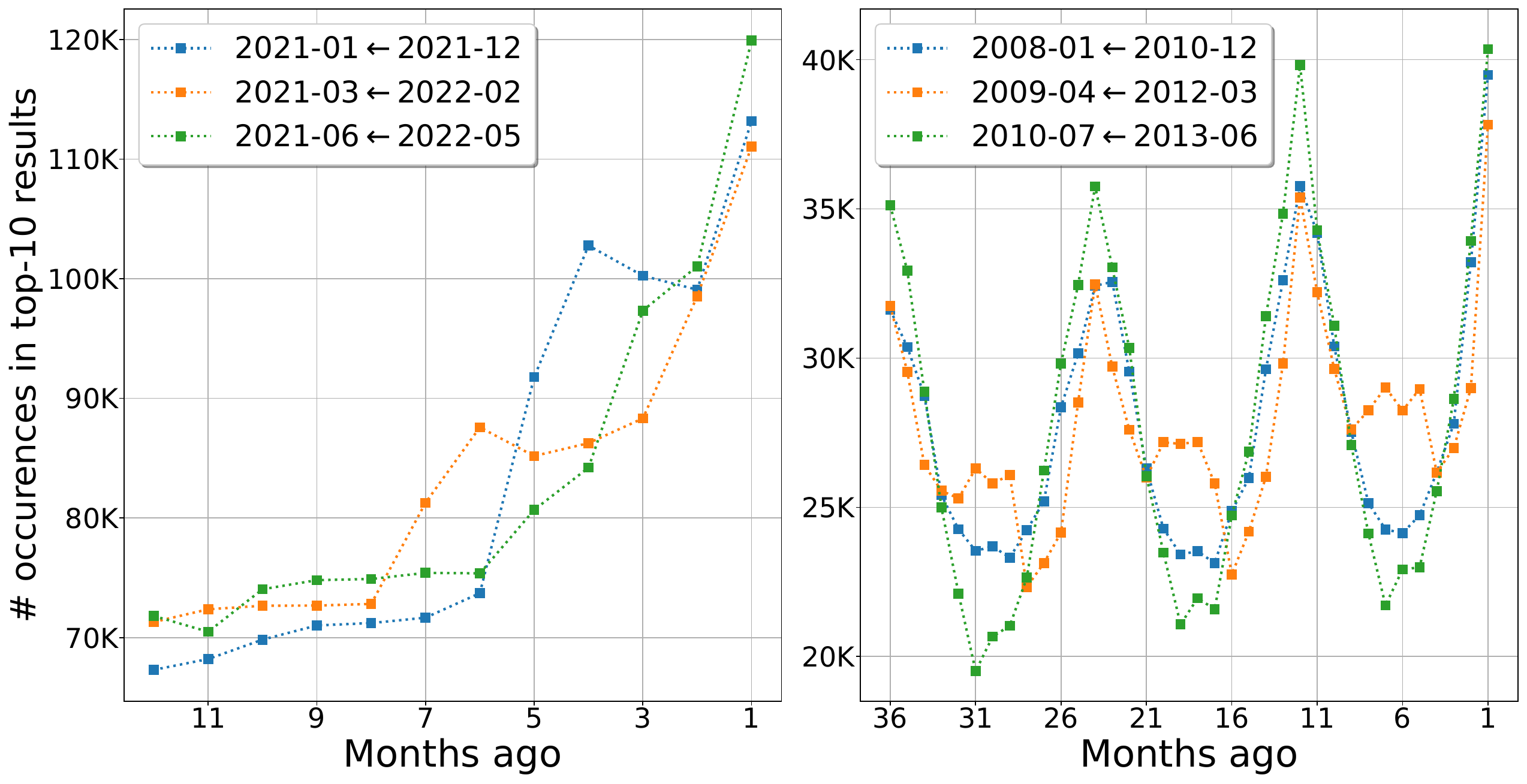}
\caption{Monthly provenance of the top-10 vectors for queries from a given month.}
\label{fig:topk_yfcc}
\end{figure}

\subsection{Temporal distribution of nearest neighbors} 

In this experiment, we investigate the distribution of the nearest neighbors results over predefined time periods. 
We consider queries $\Phi_i$ for month $i$ and search exact top-k (k${=}10$) results in $\Phi_{i-1}\cup ... \cup \Phi_{i-m}$ ($m{=}12$ for VideoAds and $m{=}36$ for YFCC). 
Figure~\ref{fig:topk_yfcc} shows how many of the $\lvert \Phi_{i} \rvert{\times}k$ nearest neighbors come from each of the $m$ previous months for a few settings of the reference month $i$. 
We observe that recent content occurs more often in the results for both datasets. 
Also, for YFCC, there is a clear seasonal trend, \ie content from the same season in years before is more likely to be returned. 
This hints at a drift because if the vectors were temporally i.i.d. search results would be equally likely to come from all  previous $m$ months.

\section{Content drift with static indexes}
\label{sect:notation}
\label{sec:indexdrift} 

In this section, we assess the effect of content drift for various index types. 

\subsection{Experimental setting}
\label{sect:setting}

The evaluation protocol follows a common use case, where the most recent content is used as query data against a backlog of older content. 

\noindent
\textbf{Data streaming.} 
The index is trained on months $\{i,...,{i+m{-}1}\}$ and months $\{j,...,{j+m{-}1}\}$ are added to the index, where $m$ is a window size.
In the \textit{in-domain} (ID) setting, the training is performed on the same data as the indexed vectors ($i{=}j$).
In the \textit{out-of-domain} (OOD) setting, the training and indexing time ranges are different ($i{=}0$, the first month of the time series and $i{\ne}j$).
If not stated otherwise, we consider $m{=}3$ months as it is the maximum retention period of user data for many sharing platforms.

Over time, the index content is updated using a \textit{sliding window} with a time step 1 month: at each month, we remove the data for the oldest month and insert the data for the next month. 
The queries are a random subset of 10,000 vectors for the month ${j+m}$. 
This setting mimics the real-world setting when queries come after the index is constructed.


\noindent
\textbf{Metrics.} 
Our ground truth is simply the exact k-NN results from brute force search. 
As a primary quality measure, we use \textit{$k{-}recall@{k}$}, the fraction of ground-truth $k$-nearest neighbors found in the top $k$ results retrieved by the search algorithm ($k{=}10$)~\cite{simhadri22results}. 
For IVF-based indexes, the efficiency measure is the number of \textit{distance calculations} (DCS) between the query vector and database entries~\cite{Babenko12}. 
At search time, the DCS is given as a fixed ``budget'':
for a query vector $q$, clusters are visited staring from the nearest centroid to $q$. 
The distances between $q$ and all vectors in the clusters are computed until the DCS budget is exhausted (only a fraction of the vectors of the last cluster may be processed). 
The advantage of this measure over direct timings is that it is independent of the hardware. 
Appendix~\ref{app:timings} reports search time equivalents for a few settings. 

\noindent
\textbf{Evaluation.}
We evaluate the k-recall@k after each content update step over the entire time range. 
For each index type and DCS setting, we focus on the month where the gap between the ID and OOD settings is the largest.




\subsection{Robustness of static indexing methods}
\label{sect:static_methods}

In this section, we investigate the robustness of static indexes of the IVF family.
Graph-based indexes~\cite{HNSW,fu2017fast} are not in the scope of this study because they don't scale as well to billion-scale datasets. 
The components of an IVF index are 
(1) the codec that determines how the vectors are encoded for in-memory storage
(2) the coarse quantizer that determines how the database vectors are clustered.

\noindent
\textbf{Vector codecs} decrease the size of the vectors for in-memory storage. 
We consider PCA dimensionality reduction to $128$ and $256$ dimensions and quantization methods
PQ and OPQ~\cite{Jegou11a,Opq13} to reduce vectors to $16$ and $32$ bytes. 
To make results independent of the IVF structure, we measure the 10-recall@10 with an exhaustive comparison between the queries and all vectors in $\mathcal{D}$. 
In \tab{robust_codecs}, we observe a small gap for most settings (${<}1\%$). 
We attribute this to the small codebook sizes of the codecs. 
In ML terms, they can hardly fit the data and, hence, may be less sensitive to the drift.

\begin{table}
\centering
\resizebox{0.32\textwidth}{!}{
\begin{tabular}{c|cc|cc}
\toprule
& \multicolumn{2}{c|}{VideoAds} & \multicolumn{2}{c}{YFCC}\\ 
Method & ID & OOD & ID & OOD \\ 
\midrule
$PCA128$ & 0.709 &\cellcolor[HTML]{d7ffb2} 0.698 & 0.625 &\cellcolor[HTML]{bcffb2} 0.622 \\
$PCA256$ & 0.844 &\cellcolor[HTML]{d1ffb2} 0.835 & 0.867 &\cellcolor[HTML]{bcffb2} 0.864 \\
\midrule
$PQ16$   & 0.237 &\cellcolor[HTML]{b2ffb2} 0.237 & 0.164 &\cellcolor[HTML]{bfffb2} 0.160 \\
$PQ32$   & 0.441 &\cellcolor[HTML]{b2ffb2} 0.441 & 0.380 &\cellcolor[HTML]{bcffb2} 0.377 \\
\midrule
$OPQ16$  & 0.457 &\cellcolor[HTML]{d7ffb2} 0.446 & 0.302 &\cellcolor[HTML]{ccffb2} 0.294 \\
$OPQ32$  & 0.609 &\cellcolor[HTML]{ccffb2} 0.601 & 0.484 &\cellcolor[HTML]{c9ffb2} 0.477 \\
\bottomrule
\end{tabular}}
\caption{
Relative 10-recall@10 for in-domain (ID) and out-of-domain (OOD) search, with various vector compression methods. 
The recall degradation is considered acceptable. 
}
\label{tab:robust_codecs}
\end{table}

\begin{table}
\resizebox{0.48\textwidth}{!}{
\setlength\tabcolsep{3.8pt}
\renewcommand{\arraystretch}{1.2}
\hspace*{-5mm}
\begin{tabular}{c|cc|cc|cc|cc}
\toprule
Budget & \multicolumn{2}{c|}{6000 DCS} & \multicolumn{2}{c|}{12000 DCS} & \multicolumn{2}{c|}{30000 DCS} & \multicolumn{2}{c}{60000 DCS}\\

Method & ID & OOD & ID & OOD & ID & OOD & ID & OOD  \\
\midrule
IVF$16384$ & 0.842 &\cellcolor[HTML]{ffb2be} 0.732 & 0.914 &\cellcolor[HTML]{ffb2be} 0.845 & 0.966 &\cellcolor[HTML]{ffefb2} 0.938 & 0.985 &\cellcolor[HTML]{dbffb2} 0.973 \\
IVF$65536$ & 0.914 &\cellcolor[HTML]{ffb2be} 0.835 & 0.956 &\cellcolor[HTML]{ffb3b2} 0.910 & 0.984 &\cellcolor[HTML]{ebffb2} 0.967 & 0.993 &\cellcolor[HTML]{ccffb2} 0.985 \\
\midrule
IMI$2{\times}8$  & 0.257 &\cellcolor[HTML]{dbffb2} 0.245 & 0.391 &\cellcolor[HTML]{fff2b2} 0.364 & 0.662 &\cellcolor[HTML]{ffb2be} 0.592 & 0.838 &\cellcolor[HTML]{ffb2be} 0.775\\
IMI$2{\times}10$ & 0.529 &\cellcolor[HTML]{ffb2be} 0.469 & 0.732 &\cellcolor[HTML]{ffb2be} 0.651 & 0.891 &\cellcolor[HTML]{ffb2be} 0.841 & 0.951 &\cellcolor[HTML]{feffb2} 0.928\\
\midrule
RCQ$10\_4$ & 0.651 &\cellcolor[HTML]{ffb2be} 0.531 & 0.776 &\cellcolor[HTML]{ffb2be} 0.676 & 0.899 &\cellcolor[HTML]{ffb2be} 0.838 & 0.951 &\cellcolor[HTML]{ffd6b2} 0.916\\
RCQ$12\_4$ & 0.809 &\cellcolor[HTML]{ffb2be} 0.713 & 0.895 &\cellcolor[HTML]{ffb2be} 0.826 & 0.958 &\cellcolor[HTML]{ffddb2} 0.925 & 0.981 &\cellcolor[HTML]{e4ffb2} 0.966\\
\bottomrule
\toprule
Budget & \multicolumn{2}{c|}{6000 DCS} & \multicolumn{2}{c|}{12000 DCS} & \multicolumn{2}{c|}{30000 DCS} & \multicolumn{2}{c}{60000 DCS}\\

Method & ID & OOD & ID & OOD & ID & OOD & ID & OOD  \\
\midrule
IVF$4096$ & 0.796 &\cellcolor[HTML]{ffb2be} 0.744 & 0.892 &\cellcolor[HTML]{ffd9b2} 0.858 & 0.960 &\cellcolor[HTML]{e4ffb2} 0.945 & 0.983 &\cellcolor[HTML]{c5ffb2} 0.977 \\
IVF$16384$ & 0.894 &\cellcolor[HTML]{ffbcb2} 0.851 & 0.947 &\cellcolor[HTML]{fff8b2} 0.922 & 0.981 &\cellcolor[HTML]{d1ffb2} 0.972 & 0.993 &\cellcolor[HTML]{bfffb2} 0.989 \\
\midrule
IMI$2{\times}6$  & 0.245 &\cellcolor[HTML]{ffefb2} 0.217 & 0.409 &\cellcolor[HTML]{ffb2bb} 0.360 & 0.727 &\cellcolor[HTML]{ffb3b2} 0.681 & 0.872 &\cellcolor[HTML]{bcffb2} 0.869\\
IMI$2{\times}8$ & 0.449 &\cellcolor[HTML]{ffe6b2} 0.419 & 0.673 &\cellcolor[HTML]{ffb3b2} 0.627 & 0.870 &\cellcolor[HTML]{feffb2} 0.847 & 0.948 &\cellcolor[HTML]{d4ffb2} 0.938\\
\midrule
RCQ$8\_4$ & 0.575 &\cellcolor[HTML]{ffb2be} 0.521 & 0.730 &\cellcolor[HTML]{ffb2b8} 0.682 & 0.878 &\cellcolor[HTML]{ffefb2} 0.850 & 0.940 &\cellcolor[HTML]{deffb2} 0.927\\
RCQ$10\_4$ & 0.768 &\cellcolor[HTML]{ffb2be} 0.718 & 0.872 &\cellcolor[HTML]{ffddb2} 0.839 & 0.949 &\cellcolor[HTML]{e8ffb2} 0.933 & 0.978 &\cellcolor[HTML]{c9ffb2} 0.971\\

\midrule 
\end{tabular}}
\caption{
Relative performance of index structures for in-domain (ID) and out-of-domain (OOD) search on VideoAds (top) and YFCC (bottom).
The color shade indicates the performance drop between ID and OOD: green is $<$1\%, red is $>$5\%, yellow in-between. 
}
\vspace{-1mm}
\label{tab:robust_indexing_structures}
\end{table}

\noindent
\textbf{The coarse quantizer} 
partitions the search space. 
We evaluate the following coarse quantizers with $K$ centroids: 
IVF$K$ is the plain k-means~\cite{Jegou11a}; 
IMI$2{\times}n$ a product quantizer with two sub-vectors~\cite{Babenko12} each quantized to $2^n$ centroids ($K=2^{2\times n}$);
and RCQ$n_1$\_$n_2$ is a two-level residual quantizer~\cite{RVQ10}, where the quantizers are of sizes $2^{n_1}$ and $2^{n_2}$ centroids ($K=2^{n_1 + n_2}$).

Table~\ref{tab:robust_indexing_structures} reports the 10-recall@10 for various DCS budgets representing different operating points. 
We experiment with different index settings for VideoAds and YFCC, due to their size difference.
The content drift has a significant impact in these experiments. 
We investigate the effect of different window sizes $m$ in Appendix~\ref{app:different_m}.

\section{Updating the index to accomodate data drift}
\label{sec:indexupdate}

In this section, we introduce \ourmethod to reduce the impact of data drift on IVF indexes over time. 
First, we address the case where vectors are partitioned by the IVF but stored exactly. 
Then we show how to handle compressed vectors. 

Our baseline methods
are the lower and upper bounds of the accuracy:
{\bf None} keeps the trained part of the quantizer untouched over the entire time span;
with {\bf Full}, the index is reconstructed at each update step.

\subsection{\ourmethod updating strategies}

  
\noindent
\textbf{\ourmethod-Split} addresses imbalance by repartitioning a few clusters. 
\ourmethod-Split collects the vectors from the $k \ll K$ largest clusters into $\mathcal{B}_1$. 
The objective is to re-assign these vectors into $k_2 > k$ new clusters, where
$k_2$ is chosen so that the average new cluster size is the median cluster size $\mu$ of the whole IVF: $k_2 = \lceil |\mathcal{L}_1| / \mu \rceil$.
To keep the total number of clusters $K$ constant, vectors from the $k_2 - k$ smallest clusters are collected into $\mathcal{B}_2$. 
We train k-means with $k_2$ centroids on $\mathcal{B}_1 \cup \mathcal{B}_2$, and replace the $k_2$ involved clusters in the index. 
Other centroids and clusters are left untouched, so the update cost is much lower than $N$.

\noindent
\textbf{\ourmethod-Lazy} updates all centroids by recomputing the centroid of the vectors assigned to each cluster.
In contrast to a full k-means iteration, the vectors are \emph{not} re-assigned after the centroid update. 
Therefore, \textbf{\ourmethod-Lazy} smoothly adapts the centroids to the shifting distribution without  a costly re-assignment operation. 
\dmitry{The similar idea was previously considered in the context of VLAD descriptors~\cite{vlad}.}

\noindent
\textbf{\ourmethod-Hybrid} combines Split and Lazy by updating  the centroids first, then splitting $k$ largest clusters.

\noindent
{\bf Discussion.}
In a non-temporal setting, if the query and database vectors were sampled uniformly from the whole time range, 
they would be i.i.d. and the optimal quantizer would be k-means (if it could reach the global optimum). 
The \ourmethod approaches are heuristic, they do not offer k-means' optimality guarantees.
However, our setting is different: 
(1) the database is incremental, so we want to avoid doing a costly global k-means and 
(2) the query vectors are the most recent ones, so the i.i.d. assumption is incorrect.
Reason (1) means that we have to fall back to heuristics to ``correct'' the index on-the-fly and (2) means that \ourmethod heuristics may actually \emph{outperform} a full re-indexing.

\subsection{\ourmethod in the compressed domain}

For billion-scale datasets, the vectors stored in the IVF are compressed with codecs, see Section~\ref{sect:static_methods}. 
However, \ourmethod-Split needs to access all original embeddings (\eg from external storage). 
Besides, \ourmethod-Lazy needs to update centroids using the original vectors. 
Reconstructed vectors could be used but this significantly degrades the performance of the \ourmethod variants (see Appendix~\ref{app:dedriftrecons}).
A workaround for \ourmethod-Lazy is to store just a subsampled set of training vectors. 
There is no such workaround for the other methods: they must store the entire database, which is an operational constraint.

\paragraph{Efficient \ourmethod-Lazy with PQ compression.}

We focus on the product quantizer (PQ), which is the most prevalent and difficult to update vector codec.
There are two ways to store a database vector $x$ in an IVF index: compress directly (Section~\ref{sect:static_methods} shows that drift does not affect this much), or store by residual~\cite{Jegou11a}, which is more accurate.
When storing $x$ by residual, the vector that gets compressed is relative to the centroid $c_i$ that $x$ is assigned to: $r=x - c_i$ is compressed to an approximation $\hat{r}$. 

\newcommand{\nsq}{M_\mathrm{PQ}}
\newcommand{\ksq}{K_\mathrm{PQ}}

The distance between a query $q$ and the compressed vector $\hat{r}$ is computed in the compressed domain, without decompressing $\hat{r}$. 
For the L2 distance, this relies on distance look-up tables that are built for every $(q, c_i)$ pair, \ie when starting to process a cluster in the index.
The look-up tables are of size $\nsq\times \ksq$ for a PQ of $\nsq$ sub-quantizers of $\ksq$ entries.
In a static index, one query gets compared to the vectors of $L$ clusters ($L$ \aka $\mathrm{nprobe}$), so look-up tables are computed $L$ times. 
The runtime cost of computing look-up tables is $L\times d\times \ksq$ FLOPs. 

In \ourmethod-Lazy, there are $m$ successive ``versions'' of $c_i$.
Computing the look-up tables with only the latest version of $c_i$ incurs an unacceptable accuracy impact, so we need to compute look-up tables for each time step. 
For this, we (1) store the history of $c_i$'s values and (2) partition the vectors within each cluster into $m$ subsets, based on the version $c_i$ that they were encoded with.

The additional index storage required for the historical centroids is in the order of $K\times (m-1)\times d$ floats. 
\dmitry{For example, for $N{=}10^{7}$, $K{=2^{16}}$, $d{=}384$, $m{=}3$ and PQ32 encoding, the historical centroids stored in float16 will consume ${\sim}16\%$ of the PQ codes.
This may be considered significant, especially for large $m$ settings.
In the future work, we anticipate addressing this limitation of our approach.}

The additional computations required for the look-up tables is $L\times d\times \ksq \times (m-1)$ FLOPs. 
For large-scale static indexes, the time to build the look-up tables is small compared to the compressed-domain distance computations.
This remains true for small values of $m$. 

Note that the coarse quantization is still performed on $K$ centroids and the residuals \wrt historical centroids are computed only for the $L$ nearest centroids.

\begin{figure}
\centering
\begin{tabular}{cc}
\hspace{2mm} YFCC {\footnotesize June 2013} & \hspace{10mm} VideoAds {\footnotesize June 2022} \\
\end{tabular}
\hspace*{-4mm}
\includegraphics[width=1.08\linewidth,trim=0 0 0 0,clip]{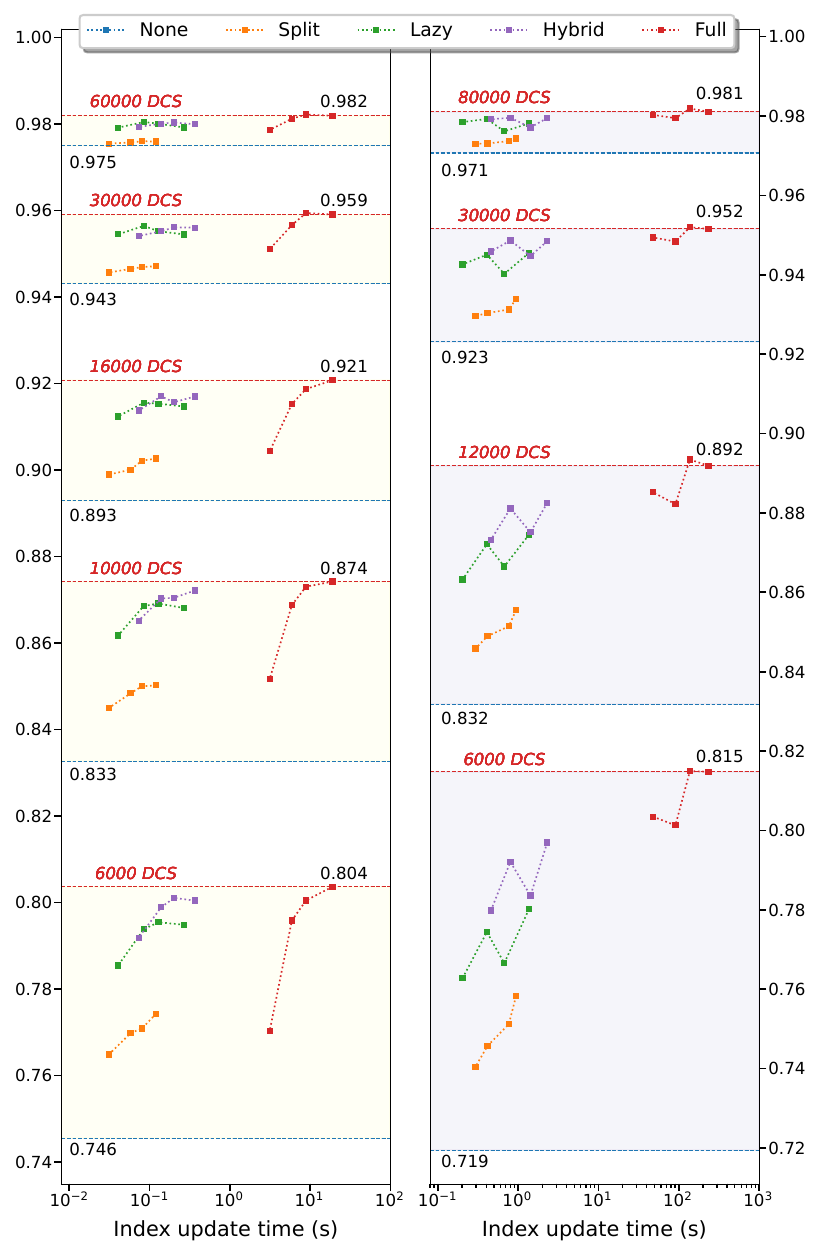} \\

\caption{\ourmethod performance tradeoff: 10-recall@10 as a function of the index update runtime. 
The upper bounds are full index reconstruction (Full).
The lower bound is no reindexing (None). 
Each point in a sequence represents an update frequency: 6, 3, 2, 1 months (left to right). 
\ourmethod variants demonstrate strong robustness to content drift on both datasets, while being two orders of magnitude faster than Full.}
\label{fig:update_methods_flat}
\end{figure}

\begin{figure*}
\centering
\vspace{5mm}
\includegraphics[width=\linewidth]{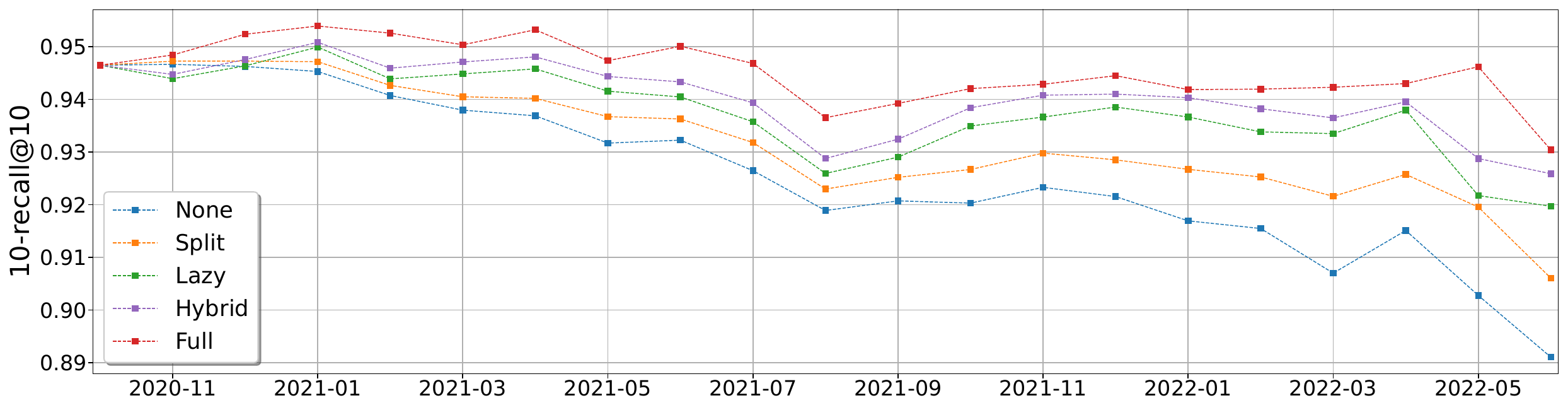} \\
\vspace{1mm}
\includegraphics[width=\linewidth]{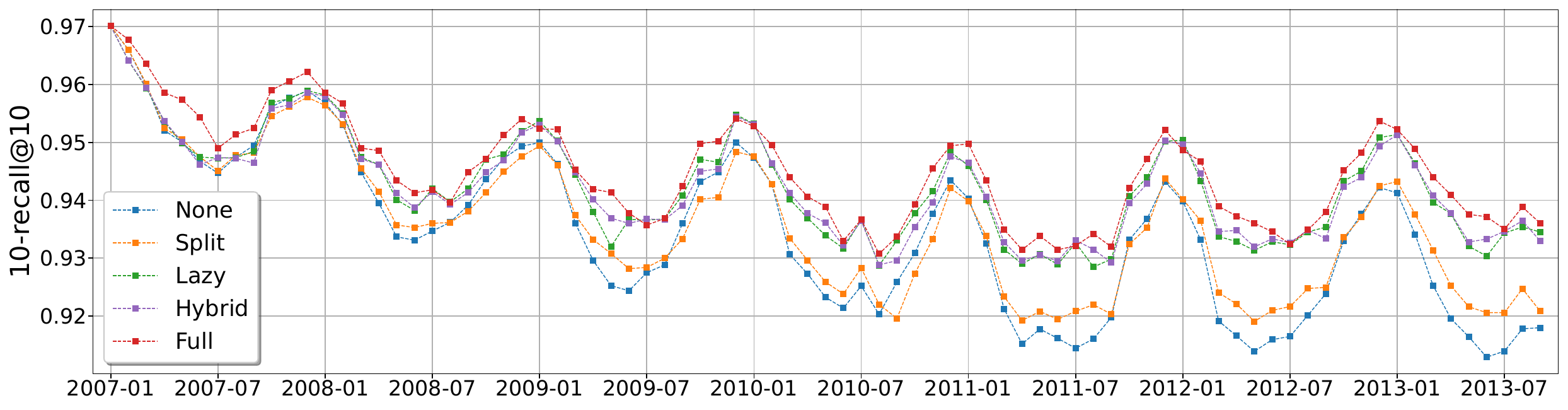} 
\caption{\ourmethod performance against None and Full over the entire time period for VideoAds (Top) and YFCC (Bottom). For both datasets, we consider $20000$ DCS budgets.
The x-axis indicates the first month of each time window ($j$).}
\label{fig:update_methods_over_time}
\end{figure*}

\section{Experimental results}
\label{sec:results}

Here, we evaluate \ourmethod empirically.
All experiments are performed with the FAISS library~\cite{johnson2017billion}, with the protocol in \sect{setting}: ANNS under a fixed budget of distance computations (DCS). 
Unless specified, we report the results at one time step $j$ chosen to maximize the average recall gap between the None and Full settings over the DCS budgets.

\begin{table}
\resizebox{0.47\textwidth}{!}{
\begin{tabular}{cccccc}
\toprule
Budget (DCS) & 6000 & 12000 & 20000 & 30000 & 60000\\
\midrule
None   &\cellcolor[HTML]{ffb2be} 0.726 &\cellcolor[HTML]{ffc9b2} 0.865 &\cellcolor[HTML]{fff8b2} 0.917 &\cellcolor[HTML]{deffb2} 0.950 &\cellcolor[HTML]{ccffb2} 0.976 \\
Split  &\cellcolor[HTML]{ffefb2} 0.796 &\cellcolor[HTML]{f4ffb2} 0.884 &\cellcolor[HTML]{e4ffb2} 0.927 &\cellcolor[HTML]{dbffb2} 0.951 &\cellcolor[HTML]{c9ffb2} 0.977 \\
Lazy   &\cellcolor[HTML]{ffb2be} 0.720 &\cellcolor[HTML]{ffb2be} 0.796 &\cellcolor[HTML]{ffb2be} 0.832 &\cellcolor[HTML]{ffb2be} 0.868 &\cellcolor[HTML]{ffcdb2} 0.946 \\
Hybrid &\cellcolor[HTML]{b2ffb2} 0.824 &\cellcolor[HTML]{bfffb2} 0.900 &\cellcolor[HTML]{c2ffb2} 0.937 &\cellcolor[HTML]{bfffb2} 0.959 &\cellcolor[HTML]{bfffb2} 0.980 \\
Full   &\cellcolor[HTML]{b2ffb2} 0.824 &\cellcolor[HTML]{b2ffb2} 0.904 &\cellcolor[HTML]{b2ffb2} 0.942 &\cellcolor[HTML]{b2ffb2} 0.963 &\cellcolor[HTML]{b2ffb2} 0.984 \\
\bottomrule
\end{tabular}}
\vspace{1mm}
\caption{\ourmethod robustness to outlier content on the YFCC dataset for IVF4096 without compression. 
\ourmethod{-}Lazy noticeably degrades when confronted with a large portion of abnormal images while \ourmethod{-}Split and \ourmethod{-}Hybrid can  successfully avoid the drop in performance. 
Numbers for $j$=September~2012.
}
\vspace{-1mm}
\label{tab:yfcc_robustness_ivfflat}
\end{table}

\subsection{Uncompressed embeddings}
\label{sect:ivfflat_results}





First, we consider \ourmethod on IVF methods with uncompressed embeddings. 
We consider the quantizer is updated every 1, 2, 3 or 6 months to see how long it can be frozen without perceptible loss in performance. 

Along with the recall, we measure the average index update time. 
The IVF contain $K=16384$ centroids for VideoAds and $K=4096$ for YFCC (accounting to the different numbers of vectors within $m{=}3$ months). 

\fig{update_methods_flat} shows that
\ourmethod{-}Split improves the search performance especially for low DCS budgets, while it is $160{\times}$ (resp. $250{\times}$) more efficient than the index reconstruction (Full) on YFCC (resp. VideoAds).
\ourmethod{-}Lazy significantly outperforms \ourmethod{-}Split on both datasets and all DCS budgets, and is still $70{\times}$ (YFCC) and $170{\times}$ (VideoAds) faster than Full.
\ourmethod{-}Hybrid further improves the \ourmethod{-}Lazy performance for low DCS. 



Overall, \ourmethod almost reaches Full on YFCC and significantly reduces the gap between Full and None on VideoAds. 
\eg, on YFCC, the proposed method provides $5.4\%$, and $1.3\%$ gains for $6000$, and $30000$ DCS budgets, respectively.
\ourmethod is about two orders of magnitude cheaper than the full index reconstruction.

In Appendix~\ref{app:evolvekmeans}, we evaluate Full with \emph{evolving k-means}~\cite{evolvekmeans} which adapts the k-means algorithm to the drift. 
This provides slightly higher recall on both datasets. 
However, the update costs are similar to the full index reconstruction.

\textbf{Hyperparameters.} 
We vary crucial parameters for the \ourmethod variants.
%
For \ourmethod{-}Split, we consider $k{=}8$ and $k{=}64$ clusters for the YFCC and VideoAds datasets, respectively. 
Higher $k$ values usually lead to slightly better recall rates at the cost of noticeably higher index update costs.
\ourmethod{-}Lazy performs a single centroid update at a time. 
Appendix \ref{app:multipleup} shows that more training iterations tend to degrade the performance.
We hypothesize that further training moves the centroids too far away from the vectors already stored to represent them accurately.

\subsection{Robustness to outlier content}
We investigate the index robustness to outlier content that occasionally occurs in real-world settings. 
When starting experiments on the YFCC dataset, we observed bursts of images of uniform color. 
It turns out these are placeholders for removed images. 
For the previous experiments, we removed these in the cleanup phase. 

To assess \ourmethod's robustness, we add these images back to the YFCC dataset and repeat the experiments from \sect{ivfflat_results}. 
\tab{yfcc_robustness_ivfflat} shows results for September 2012 (month with the largest None-Full gap):
\ourmethod{-}Lazy significantly degrades compared to all methods, \emph{including no reindexing} (None). 
In contrast, \ourmethod{-}Split and \ourmethod{-}Hybrid prevent the performance drop, Hybrid is comparable to the full index update (Full).
This shows \ourmethod{-}Split makes indexes robust to abnormal traffic. 

\begin{table}
\resizebox{0.47\textwidth}{!}{
\begin{tabular}{cccccc}
\toprule
& \multicolumn{5}{c}{IVF16384,OPQ32, direct encoding} \\
\midrule
Budget (DCS) & 6000 & 12000 & 20000 & 30000 & 60000\\
\midrule
None   &\cellcolor[HTML]{ffcccd} 0.501 &\cellcolor[HTML]{fff8cc} 0.547 &\cellcolor[HTML]{f1ffcc} 0.566 &\cellcolor[HTML]{e4ffcc} 0.577 &\cellcolor[HTML]{dbffcc} 0.586 \\
Split  &\cellcolor[HTML]{fff4cc} 0.520 &\cellcolor[HTML]{f1ffcc} 0.556 &\cellcolor[HTML]{e7ffcc} 0.571 &\cellcolor[HTML]{e0ffcc} 0.579 &\cellcolor[HTML]{d8ffcc} 0.587 \\
Lazy   &\cellcolor[HTML]{f4ffcc} 0.530 &\cellcolor[HTML]{e2ffcc} 0.563 &\cellcolor[HTML]{d8ffcc} 0.577 &\cellcolor[HTML]{d6ffcc} 0.583 &\cellcolor[HTML]{d6ffcc} 0.588 \\
Hybrid &\cellcolor[HTML]{e9ffcc} 0.535 &\cellcolor[HTML]{e0ffcc} 0.564 &\cellcolor[HTML]{dbffcc} 0.576 &\cellcolor[HTML]{d8ffcc} 0.582 &\cellcolor[HTML]{d4ffcc} 0.589 \\
Full   &\cellcolor[HTML]{ccffcc} 0.548 &\cellcolor[HTML]{ccffcc} 0.573 &\cellcolor[HTML]{ccffcc} 0.583 &\cellcolor[HTML]{ccffcc} 0.588 &\cellcolor[HTML]{ccffcc} 0.593 \\
\midrule
& \multicolumn{5}{c}{IVF16384,OPQ32, residual encoding} \\
\midrule
Budget (DCS) & 6000 & 12000 & 20000 & 30000 & 60000\\
\midrule
None   &\cellcolor[HTML]{ffccd4} 0.522 &\cellcolor[HTML]{ffcccc} 0.569 &\cellcolor[HTML]{ffe2cc} 0.589 &\cellcolor[HTML]{ffeacc} 0.599 &\cellcolor[HTML]{fff6cc} 0.608 \\
Split  &\cellcolor[HTML]{ffd2cc} 0.544 &\cellcolor[HTML]{ffe8cc} 0.582 &\cellcolor[HTML]{fff4cc} 0.597 &\cellcolor[HTML]{fff8cc} 0.605 &\cellcolor[HTML]{fcffcc} 0.613 \\
Lazy   &\cellcolor[HTML]{fff4cc} 0.559 &\cellcolor[HTML]{fcffcc} 0.593 &\cellcolor[HTML]{f4ffcc} 0.607 &\cellcolor[HTML]{f4ffcc} 0.613 &\cellcolor[HTML]{efffcc} 0.619 \\
Hybrid &\cellcolor[HTML]{fff8cc} 0.561 &\cellcolor[HTML]{fffdcc} 0.591 &\cellcolor[HTML]{faffcc} 0.604 &\cellcolor[HTML]{faffcc} 0.610 &\cellcolor[HTML]{f6ffcc} 0.616 \\
Full   &\cellcolor[HTML]{ccffcc} 0.587 &\cellcolor[HTML]{ccffcc} 0.615 &\cellcolor[HTML]{ccffcc} 0.625 &\cellcolor[HTML]{ccffcc} 0.631 &\cellcolor[HTML]{ccffcc} 0.635 \\
\bottomrule
\end{tabular}}
\vspace{1mm}
\caption{Comparison of the index update methods on the VideoAds dataset for June 2022.}
\label{tab:video_ads_ivfpq_results}
\end{table}

\subsection{PQ compressed embeddings}

We evaluate indexes with PQ compression.
We consider OPQ~\cite{Opq13} with $32$ bytes per vector.
We evaluate two settings: 
quantize either original embeddings (\textbf{direct encoding}) or their residuals w.r.t. the nearest centroid in the coarse quantizer (\textbf{residual encoding}).
Results for the VideoAds dataset are in \tab{video_ads_ivfpq_results} (see Appendix~\ref{app:results_pq_compressed} for YFCC).

The ``residual encoding'' is more sensitive to the content drift. 
Notably, \ourmethod{-}Lazy demonstrates significant improvements over no reindexing: $+3.7\%$ and $+1.4\%$ absolute for $6000$ and $30000$ DCS budgets, respectively. 
\ourmethod{-}Split also outperforms None but the gains are less pronounced compared to \ourmethod{-}Lazy.
\ourmethod{-}Hybrid does not boost \ourmethod{-}Lazy further in most cases.

\textbf{Discussion.}
\ourmethod significantly reduces the gap between full index reconstruction and doing nothing.
\ourmethod{-}Lazy is a key component that brings the most value.
We consider it as the primary technique.
\ourmethod{-}Hybrid demonstrates that \ourmethod{-}Split can be complementary to \ourmethod{-}Lazy and boost the index performance for low DCS budgets even further.
Moreover, the Split variant offers a level of robustness against sudden changes in the dataset distribution.


\section{Conclusion}
\label{sect:conclusion}

In this paper, we address the robustness of nearest neighbor search to temporal distribution drift. 
We introduce benchmarks on a few realistic and large-scale datasets, simulate the real-world settings and explore how indexing solutions degrade under drift. 
We design \ourmethod, a family of adaptations to the similarity search indexes that mitigate the content drift problem and show their effectiveness.



{\small
\bibliographystyle{ieee_fullname}
\bibliography{egbib,extra}
}

\clearpage

\newpage




\appendix 

{\LARGE \bf Appendix\\[0.1em]}

\section{Additional similarity matrices}
\label{app:simmat}

\begin{figure*}
\centering
\vspace{-2mm}
VideoAds \\
\includegraphics[width=0.95\linewidth]{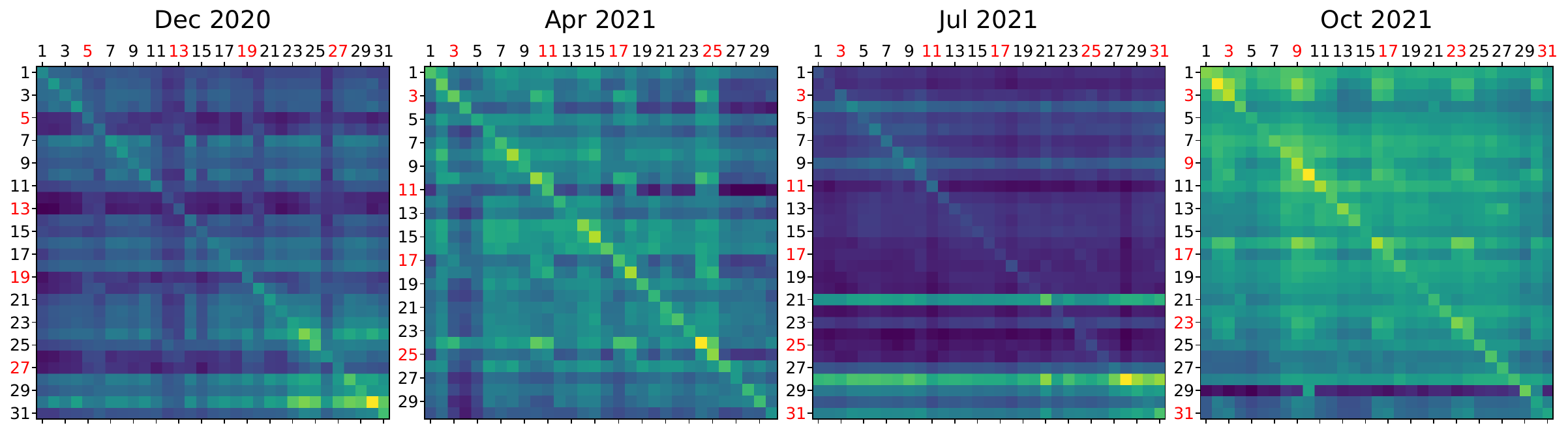}
YFCC  \\ 
\includegraphics[width=0.95\linewidth]{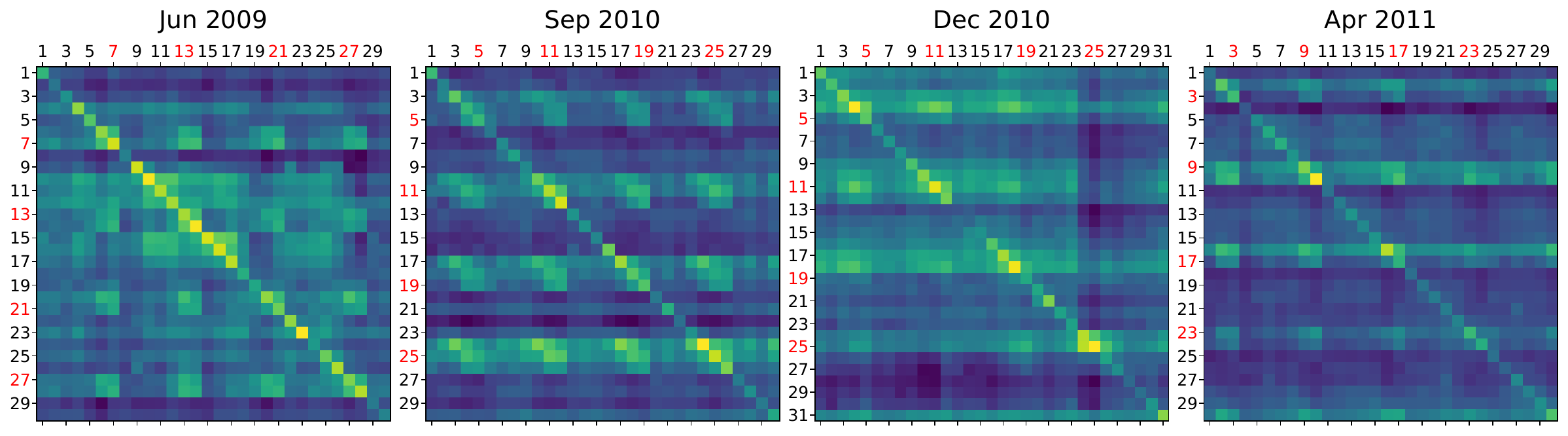}
\vspace{-1mm}
\caption{Pairwise similarity matrices between the embeddings over one month subdivided in days for a few months selected at random.
Blue and yellow correspond to low and high similarities, respectively. 
Red dates represent weekends.
Both datasets have noticeable weekday vs weekend pattern.
}
\label{fig:daily_similarity_matrices}
\vspace{-1mm}
\end{figure*}

In \fig{daily_similarity_matrices}, we provide daily similarity matrices over one month. 
We observe that the content for weekends and weekdays might differ for both datasets.
Note that the number of points per day is equalized to avoid artifacts due to the number of vectors per day.

\begin{figure}
\centering
\begin{tabular}{ccc}
\hspace{-2mm} VideoAds & \hspace{-4mm} YFCC  \\
\hspace{-2mm}{\footnotesize Jan 2021 $\rightarrow$ Dec 2021} & \hspace{-3mm}{\footnotesize Jan 2011 $\rightarrow$ Dec 2011}\\
\hspace{-3mm} \includegraphics[width=0.5\linewidth]{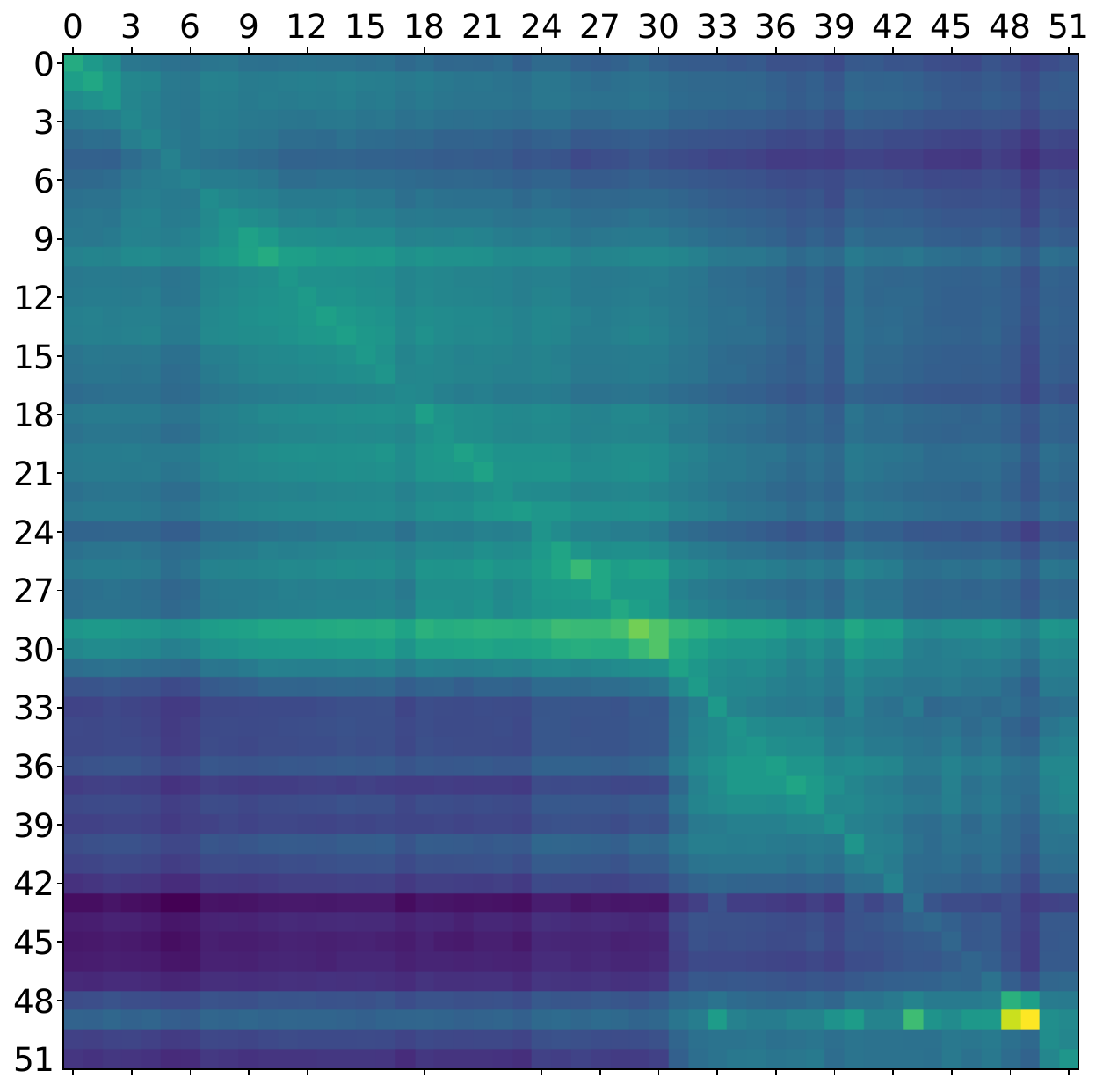} &
\hspace{-5mm}  \includegraphics[width=0.5\linewidth]{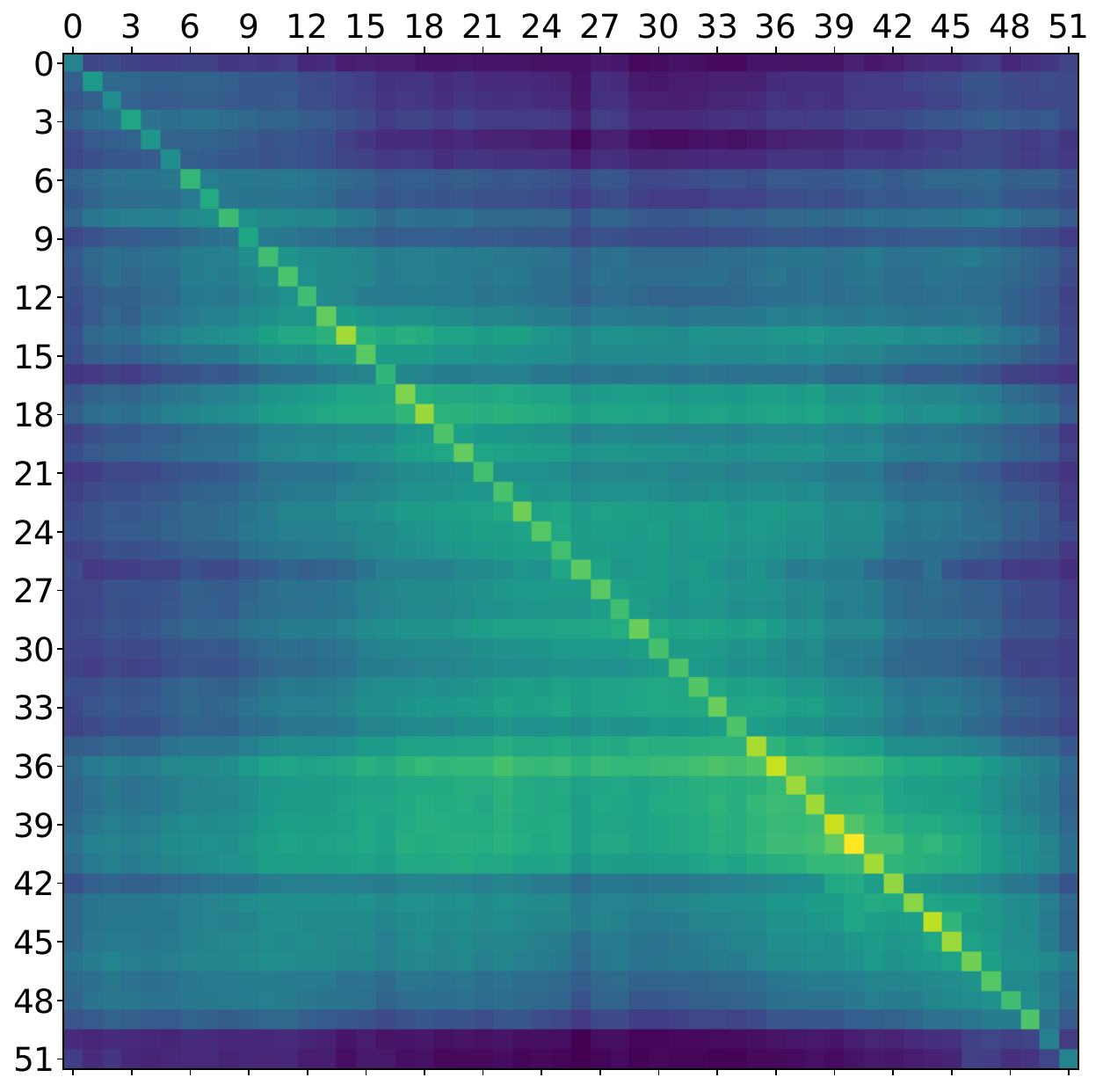} 
\end{tabular}
\caption{
    Pairwise similarity matrices between the embeddings over one year subdivided in one week.
    Blue and yellow correspond to low and high similarities, respectively.  
    There is still the seasonal pattern for YFCC and content drift over time for VideoAds.
    Both datasets do not have any clearly visible weekly correlations. 
}
\label{fig:weekly_similarity_matrices}
\vspace{1mm}
\end{figure}

\fig{weekly_similarity_matrices} presents weekly similarity matrices over one year. 
We observe the similar content drift behavior to \fig{similarity_matrices}, but no discernable weekly correlations. 

\section{Balance of K-means clusters}  
\label{app:clusterbalance}

\begin{figure}
\centering
\begin{tabular}{ccc}
\hspace{-2mm} VideoAds & \hspace{-4mm} YFCC  \\
\hspace{-2mm}{\footnotesize Oct 2020 $\rightarrow$ Sep 2022} & \hspace{-3mm}{\footnotesize Jan 2007 $\rightarrow$ Dec 2013}\\
\hspace{-3mm} \includegraphics[width=0.5\linewidth]{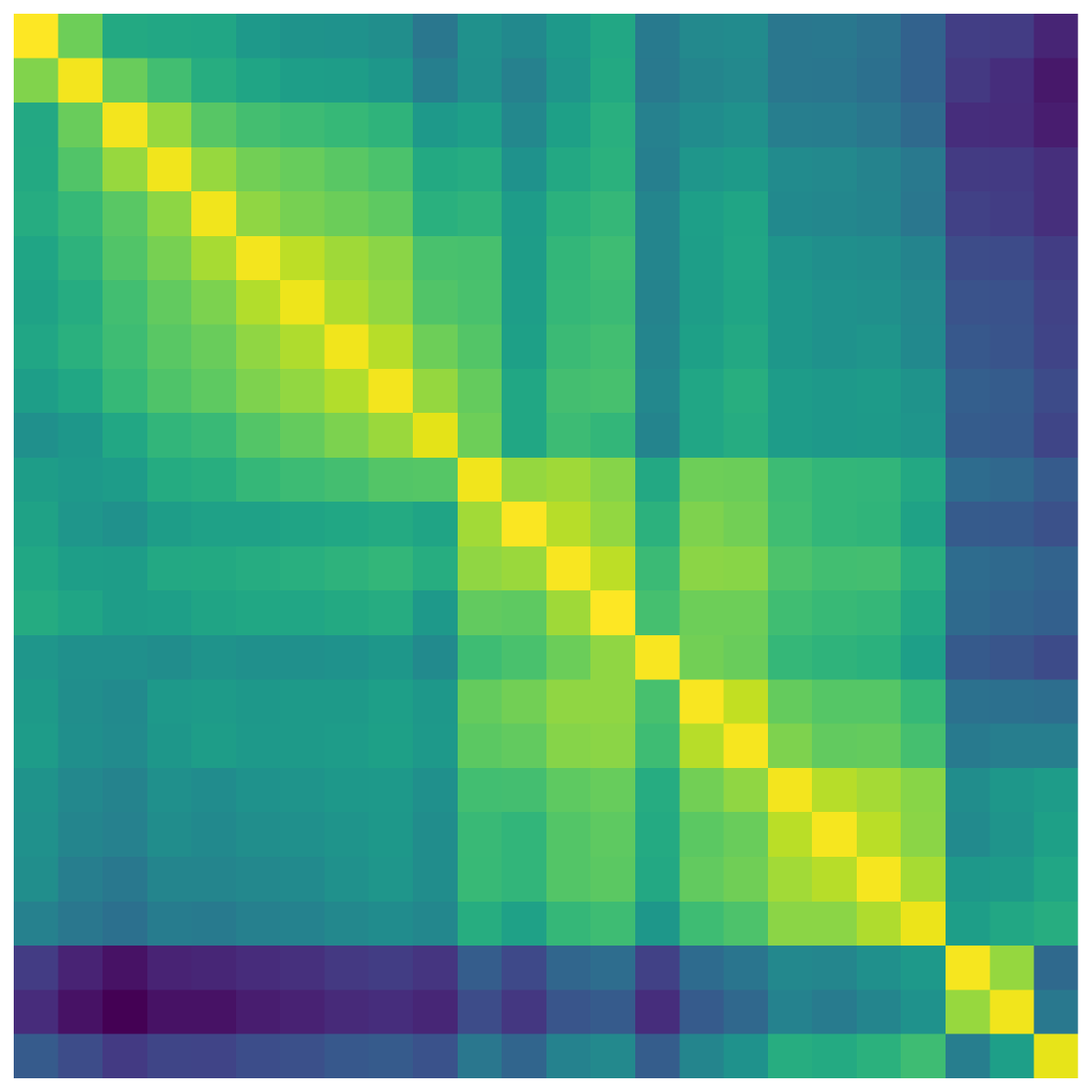} &
\hspace{-5mm}  \includegraphics[width=0.5\linewidth]{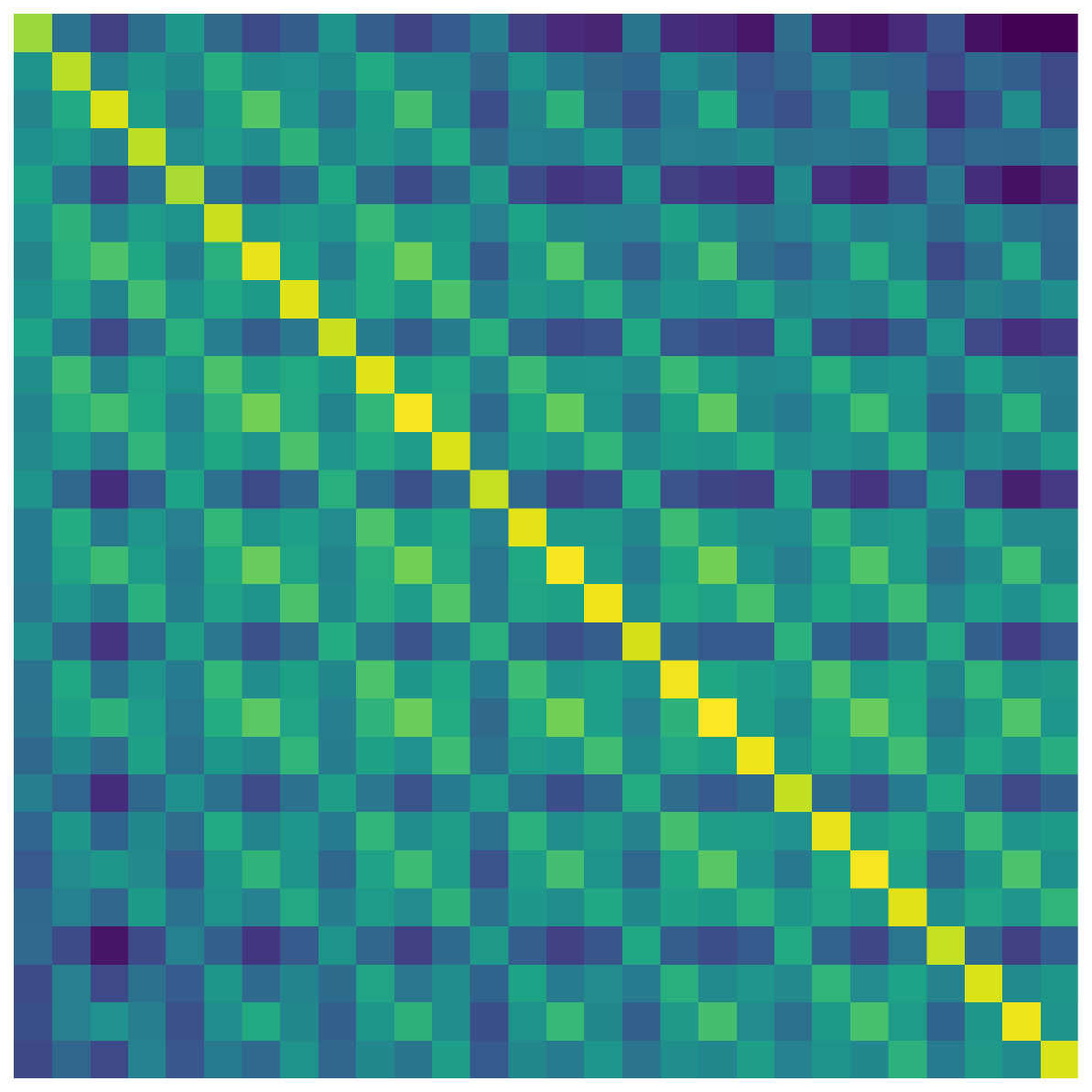} 
\end{tabular}
\caption{
    Balancedness of K-means clusters over time. 
    The starting and ending date for the periods are indicated on top.    
    For both datasets, the clusters become more imbalanced. 
    YFCC also demonstrates the seasonal behavior --- the clusters are more balanced for the same seasons than for the opposite ones. 
    Note that we use stride $3$ months for the YFCC dataset for better visualization.
}
\label{fig:balancedness}
\end{figure}

The IVF-based indexing relies on a vector quantizer to partition the vectors into clusters. 
Therefore, we investigate how content drift affects K-means clusters. 
We select months $i$ and $j$ and train K-means ($K{=}16384$) on $\Phi_i$.
Then, we assign the vectors from $\Phi_j$ to the trained centroids, count the number of points within each cluster and normalize them by $|\Phi_j|=M$. 
This yields a discrete distribution $p^{i,j}=(p_1, \dots, p_K)$ 
We use the entropy of $H(p^{i,j})$ to measure the balancedness of the K-means clusters.
For balanced clusters the entropy is $\log_2 K=14$ and for a hopelessly unbalanced clustering where all vectors are assigned to one cluster it is 0. 
Figure~\ref{fig:balancedness} shows the matrix of entropies for all pairs $(i,j)$. 
The further away from the diagonal, the lower the entropy. 
This means that the K-means clustering becomes progressively less balanced when month $i$ is more distant from month $j$. 
In addition, for YFCC, the clusters are more imbalanced for opposite seasons. 

This means that the direct distance measurements in figure~\ref{fig:similarity_matrices} translate to sub-optimal clustering as well.
For all datasets, the content drift takes place and has different nature and behavior. 
The changing distribution also affects K-means clusters and hence might lead to the noticeable degradation of the most prevalent indexing schemes at scale.

\begin{table}[h!]
\centering
\resizebox{0.48\textwidth}{!}{
\setlength\tabcolsep{3.8pt}
\renewcommand{\arraystretch}{1.2}
\hspace*{-3mm}
\begin{tabular}{cc|cc|cc|cc|cc}
\toprule
\multicolumn{2}{c|}{Budget} & \multicolumn{2}{c|}{6000 DCS} & \multicolumn{2}{c|}{12000 DCS} & \multicolumn{2}{c|}{30000 DCS} & \multicolumn{2}{c}{60000 DCS}\\
Method & $m$ & ID & OOD & ID & OOD & ID & OOD & ID & OOD  \\
\midrule

IVF$8192$ & 1 & 0.873 &\cellcolor[HTML]{ffb2be} 0.767 & 0.934 &\cellcolor[HTML]{ffb2be} 0.867 & 0.977 &\cellcolor[HTML]{ffefb2} 0.949 & 0.990 &\cellcolor[HTML]{d7ffb2} 0.979 \\
IVF$16384$ & 3 & 0.842 &\cellcolor[HTML]{ffb2be} 0.732 & 0.914 &\cellcolor[HTML]{ffb2be} 0.845 & 0.966 &\cellcolor[HTML]{ffefb2} 0.938 & 0.985 &\cellcolor[HTML]{dbffb2} 0.973 \\
IVF$32768$ & 6 & 0.839 &\cellcolor[HTML]{ffb2be} 0.738 & 0.896 &\cellcolor[HTML]{ffb2be} 0.832 & 0.956 &\cellcolor[HTML]{fff5b2} 0.930 & 0.979 &\cellcolor[HTML]{dbffb2} 0.967 \\
IVF$65536$ & 12 & 0.821 &\cellcolor[HTML]{ffb2be} 0.743 & 0.896 &\cellcolor[HTML]{ffb3b2} 0.850 & 0.955 &\cellcolor[HTML]{eeffb2} 0.937 & 0.978 &\cellcolor[HTML]{d1ffb2} 0.969 \\

\midrule
\multicolumn{2}{c|}{Budget} & \multicolumn{2}{c|}{6000 DCS} & \multicolumn{2}{c|}{12000 DCS} & \multicolumn{2}{c|}{30000 DCS} & \multicolumn{2}{c}{60000 DCS}\\
Method & $m$ & ID & OOD & ID & OOD & ID & OOD & ID & OOD  \\
\midrule

IVF$2048$ & 1 & 0.876 &\cellcolor[HTML]{ffb2be} 0.826 & 0.938 &\cellcolor[HTML]{fff5b2} 0.912 & 0.980 &\cellcolor[HTML]{d4ffb2} 0.970 & 0.992 &\cellcolor[HTML]{bcffb2} 0.989 \\
IVF$4096$ & 3 & 0.796 &\cellcolor[HTML]{ffb2be} 0.744 & 0.892 &\cellcolor[HTML]{ffd9b2} 0.858 & 0.960 &\cellcolor[HTML]{e4ffb2} 0.945 & 0.983 &\cellcolor[HTML]{c5ffb2} 0.977 \\
IVF$8192$ & 6 & 0.768 &\cellcolor[HTML]{ffb2be} 0.713 & 0.872 &\cellcolor[HTML]{ffddb2} 0.839 & 0.943 &\cellcolor[HTML]{e4ffb2} 0.928 & 0.974 &\cellcolor[HTML]{c9ffb2} 0.967 \\
IVF$16384$ & 12 & 0.758 &\cellcolor[HTML]{ffb2be} 0.703 & 0.859 &\cellcolor[HTML]{ffd3b2} 0.823 & 0.939 &\cellcolor[HTML]{e4ffb2} 0.924 & 0.973 &\cellcolor[HTML]{d1ffb2} 0.964 \\
\bottomrule
\end{tabular}}
\vspace{1mm}
\caption{Relative performance of IVF indexing structures for in-domain
(ID) and out-of-domain (OOD) search on VideoAds (top) and
YFCC (bottom) for different window sizes $m$ in months. The search accuracy measure is 10{-}recall@10. The drops in performance are essentially similar for various $m$ settings.}
\label{tab:different_m}
\vspace{-6mm}
\end{table}

\section{Robustness of indexing structures for different window sizes}
\label{app:different_m}
In our experiments, we consider the window size $m{=}3$ months which is motivated by the reasonable practical scenario. 
However, one can consider different $m$ settings.

In \tab{different_m}, we provide the robustness results for IVF indexes built upon uncompressed embeddings for various window sizes $m$ in months.  
We select the coarse quantizer sizes according to the number of datapoints within the index.
We observe that the performance degradation does not differ much, even for large $m$.

\section{\ourmethod{-}Lazy with multiple training iterations}
\label{app:multipleup}

\ourmethod{-}Lazy can be considered as a warm-started k-means to adapt to the new data distribution. 
Therefore, we investigate the impact of the number of centroid update steps $L$. 
For a normal k-means clustering the number of iterations strikes a tradeoff between speed and the quality of the clustering. 
However, \tab{lazy_iters_results} demonstrates that a single centroid update provides the highest recall.
Moreover, the number of training iterations $L{>}2$ leads to noticeable degradation.
This is because \ourmethod{-}Lazy do not reassign the points after the centroid update and hence more iterations imply that the centroids move far away from the ones that the ``old'' vectors were assigned to.
Therefore, it is both more efficient and more accurate to do a single centroid update step.

\section{Index update costs for IVF with PQ compressed embeddings on YFCC}
\label{app:costs_pq_compressed}
\tab{ivfpq_update_costs} provides the update costs for the IVF index with OPQ encoding.
On both datasets, \ourmethod demonstrates efficiency gains from 3$\times$ to 10$\times$. 

Note that the gains are smaller than for IVF operating on uncompressed embeddings.
This is because, in this experiment, the index on the PQ compressed vectors uses original data on the disk and loads it into RAM at each update step. 
This is an implementation choice, that in addition makes the timings dependent on the performance of the external storage.
Specifically, in our case, the data loading takes ${\sim}$1.7s and ${\sim}$8s for YFCC and VideoAds, respectively.


\begin{table}
\resizebox{0.47\textwidth}{!}{
\begin{tabular}{cccccc}
\toprule
\multicolumn{6}{c}{YFCC, IVF4096,Flat, Jun 2013} \\
\midrule
Budget (DCS) & 6000  & 12000 & 20000 & 30000 & 60000\\
\midrule
$L{=}0$  &\cellcolor[HTML]{ffb2bb} 0.746 &\cellcolor[HTML]{ffe3b2} 0.858 &\cellcolor[HTML]{ebffb2} 0.913 &\cellcolor[HTML]{d7ffb2} 0.943 &\cellcolor[HTML]{bfffb2} 0.975 \\
$L{=}1$  &\cellcolor[HTML]{b2ffb2} 0.795 &\cellcolor[HTML]{b2ffb2} 0.889 &\cellcolor[HTML]{b2ffb2} 0.930 &\cellcolor[HTML]{b2ffb2} 0.954 &\cellcolor[HTML]{b2ffb2} 0.979 \\
$L{=}2$  &\cellcolor[HTML]{b2ffb2} 0.795 &\cellcolor[HTML]{b5ffb2} 0.888 &\cellcolor[HTML]{b2ffb5} 0.931 &\cellcolor[HTML]{b2ffb2} 0.954 &\cellcolor[HTML]{b2ffb2} 0.979 \\
$L{=}3$  &\cellcolor[HTML]{bfffb2} 0.791 &\cellcolor[HTML]{c2ffb2} 0.884 &\cellcolor[HTML]{b8ffb2} 0.928 &\cellcolor[HTML]{b8ffb2} 0.952 &\cellcolor[HTML]{b5ffb2} 0.978 \\
$L{=}5$  &\cellcolor[HTML]{d4ffb2} 0.785 &\cellcolor[HTML]{d4ffb2} 0.879 &\cellcolor[HTML]{c5ffb2} 0.924 &\cellcolor[HTML]{c2ffb2} 0.949 &\cellcolor[HTML]{bcffb2} 0.976 \\
$L{=}10$ &\cellcolor[HTML]{eeffb2} 0.777 &\cellcolor[HTML]{eeffb2} 0.871 &\cellcolor[HTML]{d7ffb2} 0.919 &\cellcolor[HTML]{d1ffb2} 0.945 &\cellcolor[HTML]{c5ffb2} 0.973 \\
\midrule
\multicolumn{6}{c}{VideoAds, IVF16384,Flat, Jun 2022} \\
\midrule
Budget (DCS) & 6000 & 12000 & 20000 & 30000 & 60000\\
\midrule
$L{=}0$  &\cellcolor[HTML]{ffb2be} 0.719 &\cellcolor[HTML]{ffbcb2} 0.832 &\cellcolor[HTML]{ffebb2} 0.891 &\cellcolor[HTML]{feffb2} 0.923 &\cellcolor[HTML]{d4ffb2} 0.961 \\
$L{=}1$  &\cellcolor[HTML]{b2ffb2} 0.780 &\cellcolor[HTML]{b2ffb2} 0.875 &\cellcolor[HTML]{b2ffb2} 0.920 &\cellcolor[HTML]{b2ffb2} 0.946 &\cellcolor[HTML]{b2ffb2} 0.971 \\
$L{=}2$  &\cellcolor[HTML]{b2ffb2} 0.780 &\cellcolor[HTML]{c5ffb2} 0.869 &\cellcolor[HTML]{c9ffb2} 0.913 &\cellcolor[HTML]{c9ffb2} 0.939 &\cellcolor[HTML]{c2ffb2} 0.966 \\
$L{=}3$  &\cellcolor[HTML]{c9ffb2} 0.773 &\cellcolor[HTML]{dbffb2} 0.863 &\cellcolor[HTML]{d7ffb2} 0.909 &\cellcolor[HTML]{dbffb2} 0.934 &\cellcolor[HTML]{d1ffb2} 0.962 \\
$L{=}5$  &\cellcolor[HTML]{d7ffb2} 0.769 &\cellcolor[HTML]{e4ffb2} 0.860 &\cellcolor[HTML]{e8ffb2} 0.904 &\cellcolor[HTML]{e8ffb2} 0.930 &\cellcolor[HTML]{deffb2} 0.958 \\
$L{=}10$ &\cellcolor[HTML]{fff2b2} 0.753 &\cellcolor[HTML]{ffe3b2} 0.844 &\cellcolor[HTML]{fff2b2} 0.893 &\cellcolor[HTML]{fff5b2} 0.920 &\cellcolor[HTML]{f4ffb2} 0.951 \\
\bottomrule
\end{tabular}}
\vspace{1mm}
\caption{
    \ourmethod{-}Lazy performance for the different number of centroid update iterations $L$. 
    $L{=}1$ provides the highest recall values.
    Note that $L{=}1$ is also the most efficient option.
}
\label{tab:lazy_iters_results}
\end{table}

\begin{table}
\centering
\resizebox{0.4\textwidth}{!}{
\begin{tabular}{ccccc}
\toprule
\multicolumn{5}{c}{YFCC, IVF4096,OPQ32} \\
\midrule
Method &  Split & Lazy & Hybrid & Full \\
Update costs (s) & 2.1 & 8.1 & 10.4 & 29.8  \\  
\midrule
\multicolumn{5}{c}{VideoAds, IVF16384,OPQ32} \\
\midrule
Method & Split & Lazy & Hybrid & Full \\
Update costs (s) & 10.8 & 24.1 & 33.2 & 430.8 \\  
\bottomrule
\end{tabular}}
\vspace{1mm}
\caption{Index update costs for IVF indexes with OPQ encoding. \ourmethod variants are much more efficient than full index reconstruction (Full).}
\label{tab:ivfpq_update_costs}
\end{table}

\section{\ourmethod on IVF with PQ compressed embeddings on YFCC}
\label{app:results_pq_compressed}

\tab{yfcc_ivfpq_results} presents the results of the IVF index with OPQ encoding  on the YFCC dataset. 
The performance drop caused by the content drift is smaller compared to VideoAds.
Nevertheless, \ourmethod almost closes the gap between no reindexing (None) and full index reconstruction (Full).

\begin{table}
\resizebox{0.47\textwidth}{!}{
\begin{tabular}{cccccc}
\toprule
& \multicolumn{5}{c}{IVF4096,OPQ32, direct encoding} \\
\midrule
Budget (DCS) & 6000 & 12000 & 20000 & 30000 & 60000\\
\midrule
None   &\cellcolor[HTML]{f8ffb2} 0.414 &\cellcolor[HTML]{d4ffb2} 0.444 &\cellcolor[HTML]{c2ffb2} 0.455 &\cellcolor[HTML]{bcffb2} 0.461 &\cellcolor[HTML]{b8ffb2} 0.465 \\
Split  &\cellcolor[HTML]{dbffb2} 0.423 &\cellcolor[HTML]{c5ffb2} 0.448 &\cellcolor[HTML]{bcffb2} 0.457 &\cellcolor[HTML]{bcffb2} 0.461 &\cellcolor[HTML]{b8ffb2} 0.465 \\
Lazy   &\cellcolor[HTML]{bcffb2} 0.432 &\cellcolor[HTML]{b8ffb2} 0.452 &\cellcolor[HTML]{b5ffb2} 0.459 &\cellcolor[HTML]{b5ffb2} 0.463 &\cellcolor[HTML]{b5ffb2} 0.466 \\
Hybrid &\cellcolor[HTML]{bcffb2} 0.432 &\cellcolor[HTML]{b5ffb2} 0.453 &\cellcolor[HTML]{b2ffb2} 0.460 &\cellcolor[HTML]{b2ffb2} 0.464 &\cellcolor[HTML]{b5ffb2} 0.466 \\
Full   &\cellcolor[HTML]{b2ffb2} 0.435 &\cellcolor[HTML]{b2ffb2} 0.454 &\cellcolor[HTML]{b2ffb2} 0.460 &\cellcolor[HTML]{b2ffb2} 0.464 &\cellcolor[HTML]{b2ffb2} 0.467 \\

\midrule
& \multicolumn{5}{c}{IVF4096,OPQ32, residual encoding} \\
\midrule
Budget (DCS) & 6000 & 12000 & 20000 & 30000 & 60000\\
\midrule
None   &\cellcolor[HTML]{fff2b2} 0.453 &\cellcolor[HTML]{f1ffb2} 0.481 &\cellcolor[HTML]{e8ffb2} 0.492 &\cellcolor[HTML]{e1ffb2} 0.497 &\cellcolor[HTML]{deffb2} 0.501 \\
Split  &\cellcolor[HTML]{ebffb2} 0.463 &\cellcolor[HTML]{deffb2} 0.487 &\cellcolor[HTML]{deffb2} 0.495 &\cellcolor[HTML]{d7ffb2} 0.500 &\cellcolor[HTML]{d7ffb2} 0.503 \\
Lazy   &\cellcolor[HTML]{c5ffb2} 0.474 &\cellcolor[HTML]{c2ffb2} 0.495 &\cellcolor[HTML]{bfffb2} 0.504 &\cellcolor[HTML]{bfffb2} 0.507 &\cellcolor[HTML]{bfffb2} 0.510 \\
Hybrid &\cellcolor[HTML]{b8ffb2} 0.478 &\cellcolor[HTML]{bfffb2} 0.496 &\cellcolor[HTML]{bfffb2} 0.504 &\cellcolor[HTML]{bfffb2} 0.507 &\cellcolor[HTML]{bfffb2} 0.510 \\
Full   &\cellcolor[HTML]{b2ffb2} 0.480 &\cellcolor[HTML]{b2ffb2} 0.500 &\cellcolor[HTML]{b2ffb2} 0.508 &\cellcolor[HTML]{b2ffb2} 0.511 &\cellcolor[HTML]{b2ffb2} 0.514 \\
\bottomrule
\end{tabular}}
\vspace{1mm}
\caption{Comparison of the index update methods on the YFCC dataset for IVF4096,OPQ32.}
\label{tab:yfcc_ivfpq_results}
\end{table}

\section{Runtimes for different budgets}
\label{app:timings}

In this section, we report measured search times in milliseconds for different DCS budgets on each dataset. 
We average the runtimes over 20 independent runs. 
All runs are performed with $30$ threads on an Intel Xeon Gold 6230R CPU @ 2.10GHz. 


\section{Running \ourmethod on reconstructed vectors}
\label{app:dedriftrecons}

In \tab{reconstructed}, we present the index update method performance if the cetroids are updated based on either original embeddings or reconstructed ones from the PQ encodings.
\ourmethod does not degrade the recall values much while the full index reconstruction is noticeably affected.

\begin{table}
\resizebox{0.48\textwidth}{!}{
\setlength\tabcolsep{3.8pt}
\renewcommand{\arraystretch}{1.2}
\hspace*{-5mm}
\begin{tabular}{c|cc|cc|cc|cc}
\toprule
\multicolumn{9}{c}{YFCC, IVF4096,OPQ32, direct encoding} \\
\midrule
Budget & \multicolumn{2}{c|}{6000 DCS} & \multicolumn{2}{c|}{12000 DCS} & \multicolumn{2}{c|}{30000 DCS} & \multicolumn{2}{c}{60000 DCS}\\
Method & Orig & Recon & Orig & Recon & Orig & Recon & Orig & Recon  \\
\midrule
Split & 0.423 &\cellcolor[HTML]{b2ffb2} 0.423 & 0.448 &\cellcolor[HTML]{b5ffb2} 0.447 & 0.461 &\cellcolor[HTML]{b2ffb2} 0.461 & 0.465 &\cellcolor[HTML]{b2ffb5} 0.466 \\
Lazy  & 0.432 &\cellcolor[HTML]{b8ffb2} 0.430 & 0.452 &\cellcolor[HTML]{b2ffb2} 0.452 & 0.463 &\cellcolor[HTML]{b2ffb2} 0.463 & 0.466 &\cellcolor[HTML]{b2ffb2} 0.466 \\
Full  & 0.435 &\cellcolor[HTML]{d4ffb2} 0.425 & 0.454 &\cellcolor[HTML]{c5ffb2} 0.448 & 0.464 &\cellcolor[HTML]{b8ffb2} 0.462 & 0.467 &\cellcolor[HTML]{b8ffb2} 0.465 \\
\toprule
\multicolumn{9}{c}{VideoAds, IVF16384,OPQ32, direct encoding} \\
\midrule
Budget & \multicolumn{2}{c|}{6000 DCS} & \multicolumn{2}{c|}{12000 DCS} & \multicolumn{2}{c|}{30000 DCS} & \multicolumn{2}{c}{60000 DCS}\\
Method & Orig & Recon & Orig & Recon & Orig & Recon & Orig & Recon  \\
\midrule 
Split & 0.520 &\cellcolor[HTML]{c5ffb2} 0.514 & 0.556 &\cellcolor[HTML]{bfffb2} 0.552 & 0.579 &\cellcolor[HTML]{b5ffb2} 0.578 & 0.587 &\cellcolor[HTML]{b5ffb2} 0.586 \\
Lazy  & 0.530 &\cellcolor[HTML]{b8ffb2} 0.528 & 0.563 &\cellcolor[HTML]{b5ffb2} 0.562 & 0.583 &\cellcolor[HTML]{b5ffb2} 0.582 & 0.588 &\cellcolor[HTML]{b2ffb2} 0.588 \\
Full  & 0.548 &\cellcolor[HTML]{f8ffb2} 0.527 & 0.573 &\cellcolor[HTML]{e1ffb2} 0.559 & 0.588 &\cellcolor[HTML]{ccffb2} 0.580 & 0.593 &\cellcolor[HTML]{c5ffb2} 0.587 \\
\bottomrule
\end{tabular}}
\vspace{1mm}
\caption{\ourmethod and full index reconstruction performance (Full) when the centroids are updated using original embeddings (Orig) and reconstructed ones from PQ encodings (Recon).}
\label{tab:reconstructed}
\end{table}


\begin{table}
\hspace{-4mm}
\resizebox{0.495\textwidth}{!}{
\setlength\tabcolsep{4.0pt}
\renewcommand{\arraystretch}{1.2}
\begin{tabular}{cccccc}
\toprule
Budget (DCS) & 6000 & 12000 & 20000 & 30000 & 60000\\
\midrule
IVF$16384$, Flat & 6.12 & 12.05 & 18.93 & 27.14 & 53.35 \\
IVF$16384$, OPQ32 & 1.08 & 1.23 & 1.29 & 1.40 & 1.96 \\
\bottomrule
\end{tabular}}
\vspace{0.1mm}
\caption{Runtimes (ms per query) for different budgets on VideoAds.}
\label{tab:runtimes_video_ads}
\end{table}

\begin{table}
\hspace{-4mm}
\resizebox{0.495\textwidth}{!}{
\setlength\tabcolsep{4.0pt}
\renewcommand{\arraystretch}{1.2}
\begin{tabular}{cccccc}
\toprule
Budget (DCS) & 6000 & 12000 & 20000 & 30000 & 60000\\
\midrule
IVF$4096$, Flat & 4.26 & 7.72 & 12.61 & 18.19 & 35.44\\
IVF$4096$, OPQ32 & 0.43 & 0.54 & 0.72 & 0.93 & 1.73\\
\bottomrule
\end{tabular}}
\vspace{0.1mm}
\caption{Runtimes (ms per query) for different budgets on YFCC.}
\label{tab:runtimes_yfcc}

\end{table}

\section{Evolving k-means evaluation}
\label{app:evolvekmeans}

In this experiment, we evaluate evolving k-means~\cite{evolvekmeans} during the full index reconstruction. 
We consider different evolving k-means configurations proposed in the paper and provide the results in \tab{evolvekmeans}.
Evolving k-means slightly improves the results on both datasets.

\begin{table}[h!]
\centering
\resizebox{0.48\textwidth}{!}{
\begin{tabular}{cccccc}
\toprule
Budget (DCS) & 6000 & 12000 & 20000 & 30000 & 60000\\
\midrule
Full naive  & 0.804 & 0.895 & 0.938 & 0.959 & 0.981 \\ 
Full~\cite{evolvekmeans} PSKV & 0.798 & 0.892 & 0.935 & 0.958 & 0.982 \\ 
Full~\cite{evolvekmeans} FSKV p{=}0.5 & 0.804 & 0.895 & 0.938 & 0.960 & 0.983 \\
Full~\cite{evolvekmeans} FSKV p{=}0.8 & \textbf{0.807} & \textbf{0.896} & \textbf{0.939} & \textbf{0.961} & \textbf{0.983} \\  
\midrule
Budget (DCS) & 6000 & 12000 & 20000 & 30000 & 60000\\
\midrule
Full naive  & 0.815 & 0.892 & 0.930 & 0.952 & 0.975 \\ 
Full~\cite{evolvekmeans} PSKV & 0.815 & 0.894 & 0.934 & 0.956 & 0.980 \\ 
Full~\cite{evolvekmeans} FSKV p{=}0.5 & 0.818 & 0.896 & 0.935 & 0.956 & 0.980 \\ 
Full~\cite{evolvekmeans} FSKV p{=}0.8 & \textbf{0.820} & \textbf{0.897} & \textbf{0.935} & \textbf{0.957} & \textbf{0.981} \\ 
\bottomrule
\end{tabular}}
\vspace{1pt}
\caption{
    Comparison with the evolving k-means method~\cite{evolvekmeans} on the YFCC (Top) and VideoAds (Bottom) datasets. 
    Evolving k-means slightly improves the recall rates for the full index reconstruction.
}
\label{tab:evolvekmeans}
\end{table}

\section{Image credits}
\label{app:imcredits}

\subsection{Attributions for Figure~\ref{fig:visialize_yfcc}}

From top to bottom and left to right, the images are from Yahoo Flickr users:

\newcommand{\monthx}[1]{\noindent {\bf #1:}}

\monthx{2007-07}
Imagine24,
fsxz,
Tuldas,
CAPow!,
Anduze traveller,
Barnkat.

\monthx{2008-10}
BEYOND BAROQUE,
armadillo444,
Anadem Chung,
nikoretro,
Jon Delorey,
Gone-Walkabout.

\monthx{2009-12}
Spider58,
thehoneybunny,
Communicore82,
Oli Dunkley,
HarshLight,
Yelp.com.

\monthx{2010-02}
cruz\_fr,
Bemep,
Dawn - Pink Chick,
ljw7189,
john.meagher,
ShashiBellamkonda.

\monthx{2011-06}
robinmyerscough,
telomi,
richmiller.photography,
hergan family,
cus73,
librarywebchic.

\end{document}